\documentclass{article}
\pdfoutput=1 

\usepackage{arxiv}

\usepackage[utf8]{inputenc} 
\usepackage[T1]{fontenc}    
\usepackage{url}            
\usepackage{booktabs}       
\usepackage{array}          
\usepackage{amsfonts}       
\usepackage{microtype}      
\usepackage{amsmath}
\usepackage{graphicx}
\usepackage{natbib}
\usepackage{float}
\usepackage{caption}
\usepackage{amssymb}
\PassOptionsToPackage{hypertexnames=false}{hyperref}
\usepackage{doi} 

\usepackage{amsthm}         
\newtheorem{assumption}{Assumption}
\newtheorem{proposition}{Proposition}
\newtheorem{remark}{Remark}
\newcommand{\E}{\mathbb{E}}

\newcommand{\logit}{\operatorname{logit}}

\title{Beyond Differences: Estimating and Tuning\\Ratio-Based Treatment Effects}

\date{November 2026}

\author{
  Michael Fuchs\thanks{Corresponding author.} \qquad Dominik Kreiss \\
  Actuarial Department \\
  Allianz Versicherungs-AG \\
  Munich, Germany \\
  \texttt{michael.fuchs@outlook.de} \quad \texttt{kreissdominik@gmail.com}
}

\renewcommand{\shorttitle}{Estimating and Tuning Ratio-Based Treatment Effects}

\hypersetup{
pdftitle={Beyond Differences: Estimating and Tuning Ratio-Based Treatment Effects},
pdfsubject={stat.ML, stat.ME},
pdfauthor={Michael Fuchs, Dominik Kreiss},
pdfkeywords={Ratio-based CATE, Q-Learner, Relative Risk, Case-Only Estimator, Hyperparameter Tuning, Uplift Modeling},
}

\providecommand{\pending}[1]{\textbf{[#1]}}

\begin{document}
\maketitle

\begin{abstract}
When treatment effects are naturally expressed as ratios --- as in
medicine, pricing, and marketing --- the ratio-based CATE
$\tau(x) = \mathbb{E}[Y|W{=}1,X{=}x] / \mathbb{E}[Y|W{=}0,X{=}x]$ is
the appropriate estimand. Yet the common estimators divide by an estimated
baseline, impose a parametric form, or regress the outcome, where a better
fit need not mean a better effect estimate. We introduce the
\emph{Q-Learner}, which writes $\tau(x)$ as a ratio of two odds, reducing
ratio-CATE estimation for binary outcomes to two propensity classification
tasks. In benchmarks on synthetic and randomized real-world data, the
Q-Learner proves especially suited to industry practice. Standard
hyperparameter tuning on its log-loss suffices to make it rank well, get the
level of the effects right and their spread more reliably than the
alternatives, whereas, for instance,
the common S-Learner needs a costly, effect-specific tuning to rank well,
often best, but then misses the spread of the effects. If the propensity is known, a single classifier fitted on only
the converters is enough. This not only speeds up development, but is often
a necessity in practice, where data are frequently collected only for
converters.
\end{abstract}


\section{Introduction}
\label{sec:introduction}

In many domains, treatment effects are naturally expressed as ratios.
In medicine, relative effects are the dominant reporting format for
both trials and subgroup analyses \citep{andersen2021}. In pricing,
price response is naturally expressed through elasticities, i.e.,
through relative rather than absolute changes in purchase rates. In
marketing, treatment effects are routinely reported as relative lift
over baseline conversion \citep{kohavi2020}. In each
case, the ratio-based CATE
\begin{equation}
\label{eq:ratio-cate}
\tau(x) = \frac{\mathbb{E}[Y|W{=}1,X{=}x]}{\mathbb{E}[Y|W{=}0,X{=}x]}
\end{equation}
is the appropriate estimand.

The ratio scale is also often more stable than the absolute scale.
\citet{schmid1998} give empirical evidence: control-group event rates
predict absolute but not relative treatment effects across clinical
trials, suggesting that the ratio scale better captures the intrinsic
treatment effect. Moreover, the same absolute effect
$\tau_{\text{diff}}(x) = 0.05$ means very different things at
different baselines: a sixfold increase when $\mu_0(x) = 0.01$ but
only a 25\% lift when $\mu_0(x) = 0.2$. The ratio scale is thus most
informative where baselines are low and vary across units, as
is common in clinical trials and marketing campaigns.

This paper grew out of exactly this low-baseline regime in industrial
practice, where we faced three challenges in estimating the ratio CATE.
First, in conversion-rate optimization and insurance pricing, baseline
purchase or claim probabilities are routinely in the low single digits
and, within fine customer segments, can approach zero. Few converters
then carry the information on the effect, and an estimator that recovers
the ratio by dividing by an estimated baseline $\hat{\mu}_0(x)$ places a
near-zero quantity in the denominator, where small absolute errors can
translate into large errors in $\hat{\tau}(x)$. Second, outcome models are hard
to tune for the effect, a well-known problem in CATE model selection
\citep{schuler2018comparison, curth2023search}: their model class and hyperparameters are
selected for predicting the outcome, and a better outcome prediction
need not mean a better effect estimate. Third, detailed data are often
collected only for the converters, while of the non-converters little more
than their number is known, so that any model fitted on all units must do
without the most informative features.

Common estimators in the CATE literature fall into three groups, none of which meets all three
challenges. The first divides by an estimated baseline.
Difference-based CATE estimation,
$\tau_{\text{diff}}(x) = \mathbb{E}[Y|W{=}1,X{=}x] - \mathbb{E}[Y|W{=}0,X{=}x]$,
is a mature area, with established meta-learners
\citep{kunzel2019metalearners, nie2021quasi} and doubly robust
constructions \citep{kennedy2023optimal}, but recovering the ratio via
$\tau(x) = 1 + \tau_{\text{diff}}(x)/\mu_0(x)$ brings the division
back, with no numerator of the same order to offset the errors in
$\hat\tau_{\text{diff}}$ and $\hat\mu_0$. The orthogonal ratio learners
developed concurrently with this work \citep{ge2026orthogonal} divide
in the same way in their correction terms. The second imposes a
parametric log-linear structure that cannot capture complex
heterogeneity \citep{yadlowsky2021ratio, zou2004modified}. The third
regresses the outcome, S- or T-Learner style
\citep{kunzel2019metalearners}, directly or on the log scale; this
avoids the uncompensated division but is tuned for the outcome, not
for the effect. Further, all of them fit their models on all units and are
therefore unsuited to the situation above, in which detailed data are
available only for the converters.

The central contribution of this paper, the
\emph{Q-Learner},\footnote{Named for the quotient-of-odds structure of
the identity in Equation~\eqref{eq:q-identity}; the Q-Learner is
unrelated to Q-learning in reinforcement learning.} meets all three challenges with a Bayes-rule identity for binary outcomes.
The identity reparameterizes the ratio CATE into two propensity
classification tasks, which any off-the-shelf classifier can solve, so that
neither a parametric form is imposed nor a baseline estimate enters a
denominator. The classifiers learn the effect rather than the outcome, so
tuning them on their log-loss, the standard workflow, tunes the effect
estimate itself. And where the propensity is known, as in randomized
experiments, a single classifier fitted on only the converters suffices.

\subsection{Contributions}

For binary outcomes (e.g., conversion or claim occurrence), where
$\tau(x)$ coincides with the conditional relative risk, this paper
makes three contributions. First, we
introduce the \emph{Q-Learner}, which exploits a Bayes-rule identity to
write the ratio CATE as a ratio of two odds:
\begin{equation*}
\tau(x) = \frac{p(x)}{1-p(x)} \cdot \frac{1-e(x)}{e(x)},
\end{equation*}
where $p(x) = P(W{=}1|Y{=}1,X{=}x)$ is the converter propensity and
$e(x) = P(W{=}1|X{=}x)$ is the treatment propensity.
This reduces ratio CATE estimation to two binary classification
tasks without requiring outcome regression. Above all, a better fit is
a better effect estimate: the log-loss of the classifier measures the
error of $\log\hat\tau(x)$ (Section~\ref{sec:q-simple}), so tuning it on
its own log-loss, the standard workflow, tunes $\hat\tau(x)$ itself. For outcome models such as the
S-Learner, this does not hold: the treatment is one weak feature among
many, and tuning toward better conversion predictions can wash out the
interactions that carry $\tau(x)$.

In randomized experiments, the known propensity can be used for $e(x)$,
which leaves a particularly simple version, the \emph{Q-Simple-Learner}: a
single classifier fitted on the converters only, known in biostatistics as
the case-only estimator \citep{vittinghoff2006case, dai2018case}
(Section~\ref{sec:related}). This makes it especially suited to industry
practice. It is computationally cheap and tuned by the
standard cross-validated log-loss, which targets $\log\hat\tau(x)$ directly
and so rewards the scale of $\hat\tau(x)$ as well as its ranking. Since it
never sees the non-converters, it works when detailed data are available
only for the converters, and it can even use features recorded only for
them, such as the chosen product or payment terms, provided the treatment
does not affect them (Section~\ref{sec:q-simple}).

Second, we derive \emph{Neyman-orthogonal augmentations} of the Q- and
Q-Simple-Learners (DR-Q, DR-Q-Simple), on both direct and log scale,
using influence-function-based corrections. For comparison, we also
augment the S- and T-Learner, paralleling concurrent work on
orthogonal learners for conditional \citep{ge2026orthogonal} and
earlier work on marginal \citep{boughdiri2025} risk ratios. These augmentations
remove the first-order bias from nuisance estimation errors. Because
the ratio is a nonlinear functional, the resulting guarantee is
weaker than the classical double robustness of difference-based
CATEs: a quadratic Taylor remainder survives
(Section~\ref{sec:dr}).

Third, we benchmark the learners on synthetic data, which vary the
conversion rate and the number of converters independently; on a
confounded variant at a fixed conversion rate, in which the propensity must
be estimated; and on seven randomized datasets from marketing, retail and
medicine (Section~\ref{sec:experiments-v2}). The number of training
converters, not the conversion rate, limits how well a ratio CATE can be
learned; a low rate mainly costs the learners that divide by an estimated
baseline their calibration, and the augmentations, without tail control, do
not improve on the plain learners. Tuning proves essential and must be
aligned with the ratio CATE: on synthetic data, the common S-Learner tuned on its log-loss
falls behind, while Qini tuning or a learner whose own log-loss targets the
effect recovers the ranking. Tuned on its log-loss, the Q(-Simple)-Learner
ranks well and gets the scale of the effects right more reliably than the
other learners, with a known propensity and, on confounded synthetic data,
an estimated one, at a
small fraction of their tuning cost; for the ranking alone, no learner wins
consistently.

The remainder of this paper is organized as follows.
Section~\ref{sec:setup} establishes notation and assumptions.
Section~\ref{sec:q-learner} derives the Q-Learner identity.
Section~\ref{sec:dr} develops the Neyman-orthogonal augmentations.
Section~\ref{sec:experiments-v2} presents the empirical evaluation on
synthetic and randomized data, with observational data in
Appendix~\ref{app:obs}, and
Section~\ref{sec:conclusion} concludes.

\subsection{Related Work}
\label{sec:related}
\paragraph{Meta-Learners and Doubly Robust Foundations.}
Difference-based CATE methodology relies on the meta-learner paradigm
of \citet{kunzel2019metalearners} and \citet{nie2021quasi}, and on the
pseudo-outcome and influence-function machinery developed in
\citet{robins1994estimation}, \citet{chernozhukov2018double},
\citet{kennedy2023optimal}, and \citet{foster2023orthogonal}. Our
DR-S/T and DR-Q constructions adapt this machinery to the ratio
functional.

\paragraph{Ratio-based CATE Estimation.}
Closest to our setting, \citet{yadlowsky2021ratio} propose a doubly
robust estimator for the ratio-based CATE restricted to a log-linear
semiparametric model, $\tau(x) = \exp(\beta_0^\top x)$. Our work imposes no
parametric structure on $\tau(x)$, introduces the Q-Learner
parameterization, which in its plug-in form requires no outcome
regression, and its orthogonal augmentations. Causal
forests \citep{athey2019grf} estimate heterogeneous effects via local
moment conditions and can in principle target non-standard functionals,
but do not provide an out-of-the-box estimator or efficiency theory for
the ratio CATE. \citet{vanderlaan2024combining} propose EP-learning for
causal contrasts including the conditional relative risk, motivated by
the instability of DR-Learners whose pseudo-outcomes violate the bounds
of the estimand --- the failure mode we observe for our augmentations in
Section~\ref{sec:experiments-v2}.

\paragraph{Relative Risk Estimation.}
Log-binomial and modified Poisson regression
\citep{zou2004modified} estimate marginal or
conditionally linear relative risks but cannot capture nonlinear
heterogeneity. \citet{richardson2017modeling} develop a semiparametric
framework for the joint modeling of the relative risk and the risk
difference, characterizing the nuisance tangent space for these
functionals. \citet{shirvaikar2023} modify causal forests to target
relative risk via GLM-based node splitting; their approach retains a
GLM-based splitting criterion and does not provide doubly robust
guarantees. \citet{boughdiri2025} derive doubly robust estimators for
the marginal risk ratio, establishing asymptotic normality and
confidence intervals; their work targets population-level effects
rather than heterogeneous CATEs. \citet{gao2025dina}
propose the DINA estimator (difference in natural parameters) via an
R-Learner-style extension for exponential families. For binary
outcomes, DINA's natural Bernoulli family targets the conditional
log-odds ratio rather than the risk ratio; for Poisson (count)
outcomes it coincides with the conditional log risk ratio, paralleling
our log-scale variants. DINA also differs in construction: it adapts
the R-Learner's Robinson-style loss via exponential-family
likelihoods, whereas our DR learners use explicit
influence-function-based pseudo-outcomes.

\paragraph{Concurrent Work.}
One closely related work appeared six weeks before the first version
of this paper: \citet{ge2026orthogonal} independently develop
orthogonal (DR- and R-Learner-style) machine learning for conditional
odds and risk ratios, deriving orthogonal risk functions whose
pseudo-outcomes satisfy second-order remainder properties analogous
to the ATE case. Their DR-Learner for the risk ratio coincides with our
DR-T augmentation, on both the direct and the log scale, which was
developed independently; the Q-Learner and the augmentations DR-Q and
DR-Q-Simple are not part of their work.

\paragraph{Case-Only and Case--Control Designs.}
The Q-Learner's central nuisance $p(x) = P(W{=}1 \mid Y{=}1, X{=}x)$
is the \emph{retrospective} view of case--control studies, where
sampling conditions on the outcome and odds-based quantities remain
identifiable \citep{prentice1979}. In biostatistics, the identity
behind the Q-Simple-Learner is known as the \emph{case-only} estimator
of treatment effects on the relative-risk scale in randomized trials
\citep{vittinghoff2006case, dai2018case, dai2019trees}; the Q-Simple-Learner
coincides with it. Its observational variant
\citep{tchetgen2010semiparametric} targets parametric interactions and
does not identify $\tau(x)$ itself. To our knowledge, this line of work
has not been connected to the CATE and meta-learner literature; we
became aware of it after the first version of this paper. Our
contributions relative to it are the extension to covariate-dependent,
estimated propensities, which identifies the full ratio CATE under
unconfoundedness (Q-Learner); the result that the case-only log-loss is aligned with the
error of $\log\tau(x)$, which makes standard tuning target the effect
(Proposition~\ref{prop:logloss}), whereas \citet{dai2019trees} note that
the classification error does not connect to the accuracy of the
estimated effects; the orthogonal augmentations; and a benchmark against outcome- and difference-based learners. The
case-only view is also why our real-data protocol may subsample the
training data by outcome (Appendix~\ref{app:protocol}).

\paragraph{Uplift Modeling and Model Selection.}
The marketing literature \citep{radcliffe2011} focuses on ranking by
treatment effect; we address both ranking and calibration. Because the
effect is never observed, selecting and tuning CATE models is itself a
known difficulty: fit on the observed outcome can be a poor guide, and
proxy criteria built from pseudo-outcomes or influence functions are
proposed instead, none of which dominates \citep{schuler2018comparison,
saito2020counterfactual, curth2023search, mahajan2024empirical}, and
benchmark results depend on the data-generating process
\citep{curth2021neglected}. The
Q-Learner needs no proxy, since the log-loss of its classifiers is
aligned with the effect; tuning the S-Learner on the Qini, as we do in
Section~\ref{sec:experiments-v2}, is one such proxy.

\section{Setup and Notation}
\label{sec:setup}

We work within the standard potential outcomes framework
\citep{rubin1974}. Each unit is characterized by a vector of pre-treatment
covariates $X \in \mathcal{X}$, a binary treatment indicator
$W \in \{0,1\}$, and a binary outcome $Y \in \{0,1\}$. 

Let $Y(1),Y(0)\in\{0,1\}$ denote the potential outcomes and assume
consistency, $Y=Y(W)$. Our target estimand is the ratio-based
conditional average treatment effect
\begin{equation}
\tau(x) = \frac{\mathbb{E}[Y(1)\mid X{=}x]}{\mathbb{E}[Y(0)\mid X{=}x]} .
\end{equation}
The ratio $\tau(x)$ quantifies how much
more likely the outcome becomes under treatment relative to control, for
units with covariates $x$.

Identification of $\tau(x)$ from observational data requires two standard
assumptions:

\begin{assumption}[Unconfoundedness]
\label{ass:unconf}
$(Y(1), Y(0)) \perp W \mid X$
\end{assumption}

\begin{assumption}[Overlap]
\label{ass:overlap}
$0 < e(x) < 1$ for all $x \in \mathcal{X}$,
\end{assumption}
where $e(x) = P(W{=}1 \mid X{=}x)$ is the propensity score. In addition,
the ratio requires a positive denominator:

\begin{assumption}[Positivity of Control Response]
\label{ass:positivity}
$\mu_0(x) := P(Y{=}1 \mid W{=}0, X{=}x) > 0$ for all $x \in \mathcal{X}$.
\end{assumption}

Under Assumptions~\ref{ass:unconf}--\ref{ass:overlap},
$\mathbb{E}[Y(w)\mid X{=}x]=\mu_w(x):=P(Y{=}1\mid W{=}w,X{=}x)$, the
response surface of arm $w$, so that $\tau(x)=\mu_1(x)/\mu_0(x)$ whenever
$\mu_0(x)>0$ (Assumption~\ref{ass:positivity}). For the results of
Section~\ref{sec:dr} we additionally require strong overlap,
$\varepsilon\le e(x)\le 1-\varepsilon$, and $\mu_0(x)\ge c>0$.

Assumption~\ref{ass:positivity} concerns the \emph{existence} of the estimand only:
$\tau(x)$ is well-defined as long as $\mu_0(x)$ is strictly positive,
however small. The practical difficulty addressed in this paper is
distinct: as $\mu_0(x) \to 0$, any estimator that places
$\hat{\mu}_0(x)$ in a denominator becomes unstable even though the
estimand itself remains perfectly well-defined. The Q-Learner
(Section~\ref{sec:q-learner}) removes this denominator; for the
estimators that retain one, we apply only a numerical underflow guard
(Section~\ref{sec:tuning-setup-v2}).

Beyond these standard quantities, the Q-Learner
(Section~\ref{sec:q-learner}) and its DR variant
(Section~\ref{sec:dr-q}) rely on two additional objects:
\begin{itemize}
    \item The \textbf{converter propensity}
    $p(x) = P(W{=}1 \mid Y{=}1, X{=}x)$: the probability of having been
    treated, conditional on a positive outcome.
    \item The \textbf{marginal conversion probability}
    $m(x) = P(Y{=}1 \mid X{=}x)$: the unconditional outcome rate given
    covariates, used by the DR-Q variant only.
\end{itemize}

\section{The Q-Learner}
\label{sec:q-learner}

We now derive a meta-learner that directly exploits the multiplicative
structure of the ratio CATE. The key observation is the following
identification result:

Under Assumptions~\ref{ass:unconf}--\ref{ass:positivity}, for binary
outcomes $Y \in \{0,1\}$:
\begin{equation}
\label{eq:q-identity}
\tau(x) = \frac{p(x)}{1-p(x)} \cdot \frac{1-e(x)}{e(x)}
\end{equation}
where $p(x) = P(W{=}1|Y{=}1,X{=}x)$ is the treatment probability among
converters and $e(x) = P(W{=}1|X{=}x)$ is the propensity. To prove the
relation, apply Bayes' theorem to
$\mu_1(x) = P(Y{=}1|W{=}1,X{=}x)$:
\begin{equation}
\mu_1(x) = \frac{P(W{=}1|Y{=}1,X{=}x) \cdot P(Y{=}1|X{=}x)}{P(W{=}1|X{=}x)}
          = \frac{p(x) \cdot m(x)}{e(x)}
\end{equation}
Similarly, $\mu_0(x) = (1-p(x)) \cdot m(x) / (1-e(x))$. Taking the
ratio, $m(x)$ cancels (it is positive, as are $1-p(x)$ and $e(x)$, by
Assumptions~\ref{ass:overlap}--\ref{ass:positivity}):
\begin{equation}
\tau(x) = \frac{\mu_1(x)}{\mu_0(x)}
         = \frac{p(x)}{1-p(x)} \cdot \frac{1-e(x)}{e(x)}
\end{equation}

The identity has an intuitive interpretation: if treatment increases
conversion ($\tau(x) > 1$), then converters are disproportionately likely
to have been treated relative to the overall propensity, i.e.,
$p(x) > e(x)$. $\tau(x)$ is thus a ratio of two odds: the odds of
treatment among converters over the odds of treatment overall. Its only
denominators, $1-p(x)$ and $e(x)$, do not involve $\mu_0(x)$. The identity
is purely distributional --- it holds on the observational joint
distribution by Bayes' rule alone; its causal reading is inherited
entirely from Assumptions~\ref{ass:unconf}--\ref{ass:overlap}, exactly as
for the outcome-regression route.

The Q-Learner identity suggests a simple estimation strategy:
\begin{enumerate}
\item \textbf{Step 1:} Estimate propensity score $\hat{e}(x)$ from the
observed treatment assignments using all observations; if $e(x)$ is
known, it can be used directly (Section~\ref{sec:q-simple})
\item \textbf{Step 2:} Estimate converter treatment probability
$\hat{p}(x)$ using only observations with $Y=1$
\item \textbf{Step 3:} Compute CATE estimate:
$\hat{\tau}(x) = \frac{\hat{p}(x)}{1-\hat{p}(x)} \cdot
\frac{1-\hat{e}(x)}{\hat{e}(x)}$
\end{enumerate}

Compared to common ratio estimators in the CATE literature, the Q-Learner offers three structural advantages:
\begin{itemize}
    \item \textbf{Tuning targets the effect}: Given a correct propensity, the log-loss of $\hat{p}$ is a
    weighted squared error of $\log\hat{\tau}$ (Section~\ref{sec:q-simple}),
    so tuning the classifier on its own log-loss tunes the effect estimate
    itself. For an
    outcome model such as the S-Learner, a better outcome prediction need not
    mean a better effect estimate
    (Sections~\ref{sec:syn-tuning-v2} and~\ref{sec:rct-v2}).
    \item \textbf{Model selection without a proxy}: Since the effect is
    never observed, selecting among CATE models usually requires proxy
    criteria (Section~\ref{sec:related}). Among candidate Q-Learners, the
    choice reduces to which fits the two propensities better, measured by
    the held-out log-loss of the classifiers, as in any classification task.
    \item \textbf{Flexibility}: Any probabilistic classifier can serve as
    base model for both stages, so that, unlike parametric approaches
    \citep{yadlowsky2021ratio, zou2004modified}, no functional form is
    imposed on $\tau(x)$.
\end{itemize}

\begin{remark}[Single Model via Offset]
\label{rem:single-model}
Taking logs of \eqref{eq:q-identity} one arrives at
\begin{equation}
    \log\tau(x) = \logit(p(x)) - \logit(e(x)).
\end{equation}
This suggests fitting a single logistic regression for $p(x)$ with
$\logit(\hat{e}(x))$ as offset:
\begin{equation}
\logit(p(x)) = \logit(\hat{e}(x)) + f(x)
\end{equation}
where $f(x)$ models the treatment effect heterogeneity. The fitted model
directly yields $\hat{\tau}(x) = \exp(\hat{f}(x))$, reducing the
two-stage procedure to a single model after propensity estimation.
This offset construction is itself classical in the case--control
literature \citep{prentice1979}, where a known sampling offset enters
a logistic model in exactly this way.
\end{remark}

\begin{remark}[Extension to Multi-Class Treatments]
The Q-Learner can be generalized to non-binary treatments by using
multi-class classification for $p(w, x) = P(W{=}w|Y{=}1, X{=}x)$ and
$e(w, x) = P(W{=}w|X{=}x)$:
\begin{equation}
    \tau(x, w_1, w_2) = \frac{p(w_1, x)}{p(w_2, x)} \cdot
    \frac{e(w_2, x)}{e(w_1, x)}
\end{equation}
The identity extends to non-negative outcomes: applying Bayes' theorem
inside the expectations gives
$\tau(x)=\frac{p_Y(x)}{1-p_Y(x)}\cdot\frac{1-e(x)}{e(x)}$, where
$p_Y(x)=\mathbb{E}[WY\mid X{=}x]/\mathbb{E}[Y\mid X{=}x]$ is the treatment
probability averaged over outcome levels with weights $y$; it is
estimated by a classifier for $W$ with sample weights $Y$, and for binary
outcomes it reduces to $p(x)$. We leave its evaluation to future work.
\end{remark}

\subsection{The Q-Simple-Learner: Learning from the Converters Alone}
\label{sec:q-simple}

In RCT settings where the propensity is known by design, it is natural
to use it directly in the Q-Learner identity~\eqref{eq:q-identity}
instead of fitting a propensity model. This leaves a single model for
$p(x)$, fitted only on the converters. We call this variant the
\emph{Q-Simple-Learner}; in biostatistics it is known as the case-only
estimator (Section~\ref{sec:related}). It has a categorical advantage over the
learners fitted on all units: since the design propensity replaces the
full-sample model, it can be applied when detailed data are available only
for the converters, as can its augmentation DR-Q-Simple
(Section~\ref{sec:dr-q-simple}). This is the natural data
layout in operational settings such as price-elasticity estimation on
existing customers, where detailed data are collected only at conversion.
The Q-Simple-Learner can then even use features recorded
only for the converters, such as the chosen product, coverage or payment
terms, provided the treatment does not affect them, which no learner fitted
on all units can use. Since such features
often carry much of the information on the effect, the richer data on the
converters can matter more than the larger sample of all units.

With a constant design propensity, its
estimate is a fixed, strictly increasing transformation of one
classifier: its
ranking depends only on how $\hat{p}$ orders the units, its
calibration only on that of $\hat{p}$, and tuning $\hat{p}$ on its
log-loss tunes $\hat{\tau}$ entirely, its scale included: the excess log-loss of
$\hat{p}$ equals a squared error of $\log\hat{\tau}$, weighted towards
regions with many converters, and a $\hat{p}$ that is calibrated among
converters gives a calibrated $\hat{\tau}$ (Appendix~\ref{app:q-simple}).
Regions with few converters carry little weight, the price of learning
from converters only. No other learner in this paper is tuned for
calibration as simply. It is also computationally cheap,
especially for large datasets. It is the learner we recommend for
randomized data when the scale matters or simplicity counts: tuned on its
log-loss, it is the best or nearly the best calibrated learner in our
benchmark and ranks well, at a small fraction of the tuning cost of the S-
and T-Learners (Section~\ref{sec:guidance-v2}).

Estimating the propensity can nevertheless help, even when it is known.
$\hat{p}(x)$ and $\hat{e}(x)$ are correlated through the same realized treatment
assignments: if a covariate region happens to contain more treated
units than expected, both $\hat{e}(x)$ and $\hat{p}(x)$ increase, and
the biases partially cancel in the product
$\frac{\hat{p}}{1-\hat{p}} \cdot \frac{1-\hat{e}}{\hat{e}}$. Using the
known $e(x)$ breaks this correlation. For average effects, this is why
an estimated propensity is asymptotically more efficient than the true
one \citep{hirano2003efficient}. In our benchmark
(Section~\ref{sec:syn-v2}), however, fitting a propensity model brings no
advantage, while the computational advantage of Q-Simple remains. In
fact, the noise of the fitted propensity model sometimes even hurts the
calibration of the Q-Learner relative to Q-Simple.

\section{Neyman-Orthogonal Augmentations}
\label{sec:dr}
The Q-Learner and the naive ratio estimator $\hat{\mu}_1(x)/\hat{\mu}_0(x)$
(S- and T-Learner) are plug-in estimators: they combine estimated nuisance
functions into $\hat{\tau}(x)$ without correction for estimation error. If any nuisance model is misspecified,
bias propagates directly into the CATE estimate.

Following the framework of \citet{kennedy2022semiparametric}, we derive
\emph{augmented} estimators that eliminate the leading-order bias through
influence-function-based corrections.\footnote{Throughout, ``influence
function'' refers to the \emph{efficient influence function} (EIF) as
established in semiparametric causal inference theory
\citep{robins1994estimation, kennedy2022semiparametric}, not to the
influence function of robust statistics, which measures the impact of
outlying observations on an estimator.} The resulting estimators are
\emph{Neyman-orthogonal}: their bias is first-order insensitive to
errors in the nuisance estimates, with only second-order remainder
terms surviving. For the nonlinear ratio functional this
orthogonality is the precise guarantee on offer --- it is weaker than
the classical double robustness familiar from difference-based (A)TE
estimation, where exact unbiasedness holds if either nuisance block is
correctly specified. The ``DR-'' prefix in the learner names (DR-S,
DR-T, DR-Q, \ldots) follows the meta-learner literature; it denotes
orthogonal de-biasing, not classical double robustness.

Concretely, for an estimand
$\tau(x)$ that depends on nuisance functions $g$ (propensity, outcome
models, etc.), the augmented pseudo-outcome takes the form
\begin{equation}
\Gamma = \hat{\tau}(x) + \hat{\phi}_{\tau}(W, Y, X; x),
\end{equation}
where $\hat{\phi}_{\tau}$ is the estimated influence-function correction
evaluated at the nuisance estimates; $\Gamma$ itself is the uncentered
efficient influence function of $\E[\tau(X)]$, since $\tau(x)$ at a
single point has none. This correction removes the
first-order (linear) bias term, leaving a conditional bias that is a
product of two nuisance errors and vanishes faster than any individual
error alone. Under sufficiently fast nuisance convergence rates
(errors of order $o_p(n^{-1/4})$), regressing the
pseudo-outcome $\Gamma$ on $X$ achieves oracle rates in the sense of
\citet{kennedy2023optimal}, and the sample average
$n^{-1}\sum_i \Gamma_i$ is a valid one-step estimator of the expected
conditional ratio $\E[\tau(X)]$
\citep{robins1994estimation, chernozhukov2018double}.
 
The ratio CATE $\tau(x) = \mu_1(x)/\mu_0(x)$ admits two natural
parameterizations: through outcome models $(\mu_1, \mu_0)$ or through
the Q-Learner reparameterization $(p, e)$. Each leads to a distinct
augmented estimator. Both are reparameterizations of the same efficient
influence function, so both are Neyman-orthogonal with respect to all
their nuisances; they differ in the second-order remainder, and hence in
which fixed misspecifications they tolerate (exact biases in
Appendix~\ref{app:verifications}).

\subsection{DR-S/T: Augmenting the Ratio via Outcome Models}

\label{sec:dr-t}
Both the S- and T-Learners estimate $\hat{\tau}(x) = \hat{\mu}_1(x)/\hat{\mu}_0(x)$ as a plug-in ratio and share the following augmentation. The
standard AIPW influence functions for the response surfaces are
$\phi_{\mu_1} = W(Y - \mu_1(x))/e(x)$ and
$\phi_{\mu_0} = (1-W)(Y - \mu_0(x))/(1-e(x))$. Applying the quotient
rule $\phi_{\tau} = (\phi_{\mu_1} - \tau \cdot \phi_{\mu_0})/\mu_0$ and
substituting nuisance estimates yields the DR-S/T pseudo-outcome:
\begin{equation}
\label{eq:dr-t}
\Gamma^{\text{DR-S/T}} = \hat{\tau}(X)
  + \frac{W(Y - \hat{\mu}_1(X))}{\hat{e}(X) \cdot \hat{\mu}_0(X)}
  - \frac{\hat{\tau}(X)(1-W)(Y - \hat{\mu}_0(X))}{(1-\hat{e}(X)) \cdot \hat{\mu}_0(X)}
\end{equation}
where $\hat{\tau}(X) = \hat{\mu}_1(X)/\hat{\mu}_0(X)$. Its robustness
properties, verified in Appendix~\ref{app:verifications}, are:
\begin{enumerate}
    \item[(a)] if the propensity model is correct
    ($\hat{e} = e$), the conditional bias is
    $(\tau - \hat{\tau})(\mu_0 - \hat{\mu}_0)/\hat{\mu}_0$ --- second-order
    in the outcome-model errors, but, due to the nonlinearity of the
    ratio, \emph{not} exactly zero, which is precisely why we
    characterize DR-S/T as Neyman-orthogonal rather than classically
    doubly robust;
    \item[(b)] if the outcome models are correct
    ($\hat{\mu}_w = \mu_w$), the estimator is conditionally unbiased
    regardless of the propensity estimate;
    \item[(c)] if both are misspecified, the bias is this term plus
    products of propensity and outcome-model errors.
\end{enumerate}
A log-scale variant follows from
$\log \tau = \log \mu_1 - \log \mu_0$ with
$\phi_{\log \mu_w} = \phi_{\mu_w}/\mu_w$:
\begin{equation}
\label{eq:dr-t-log}
\Gamma^{\text{DR-S/T}}_{\log} = \log \hat{\mu}_1(X) - \log \hat{\mu}_0(X)
  + \frac{W(Y - \hat{\mu}_1(X))}{\hat{e}(X) \hat{\mu}_1(X)}
  - \frac{(1-W)(Y - \hat{\mu}_0(X))}{(1-\hat{e}(X)) \hat{\mu}_0(X)}
\end{equation}
It can stabilize variance when effects span a wide range and inherits
the same orthogonality properties, including the quadratic remainder.

\subsection{DR-Q: Augmenting the Q-Learner}
\label{sec:dr-q}

The Q-Learner identity~\eqref{eq:q-identity} expresses $\tau(x)$ as a
product $A(x) \cdot B(x)$ with $A = p/(1-p)$ and $B = (1-e)/e$.
Applying the product rule $\phi_\tau = B\phi_A + A\phi_B$ with
$\phi_A = Y(W-p(x))/(m(x)(1-p(x))^2)$ (the converter-subpopulation
influence function lifted to the full population via $m(x)$) and
$\phi_B = -(W-e(x))/e(x)^2$ yields:
\begin{equation}
\label{eq:dr-q}
\Gamma^{\text{DR-Q}} = \hat{\tau}(X) + \frac{(1-\hat{e}(X)) \cdot Y(W-\hat{p}(X))}{\hat{e}(X) \cdot \hat{m}(X)(1-\hat{p}(X))^2} - \frac{\hat{p}(X)(W-\hat{e}(X))}{(1-\hat{p}(X))\hat{e}(X)^2}
\end{equation}
where $\hat{\tau}(X) = \frac{\hat{p}(X)}{1-\hat{p}(X)} \cdot \frac{1-\hat{e}(X)}{\hat{e}(X)}$ is the plug-in Q-Learner estimate.
At the \emph{true} nuisance values, this augmentation coincides
algebraically with that of DR-S/T~\eqref{eq:dr-t} under the identities
$\mu_1 = pm/e$ and $\mu_0 = (1-p)m/(1-e)$. At \emph{estimated}
nuisances, however, the two parameterizations are no longer
equivalent: separately fitted $(\hat{p}, \hat{e}, \hat{m})$ do not
correspond to separately fitted $(\hat{\mu}_1, \hat{\mu}_0, \hat{e})$,
and estimation errors propagate differently through the two
parameterizations. This is where the two remainders differ. DR-Q is
conditionally unbiased if $\hat{p}$ and $\hat{e}$ are correct, whatever
$\hat{m}$; with $\hat{e} = e$ and $\hat{m} = m$, its bias is
second-order in $\hat{p} - p$. A misspecified $\hat{e}$ leaves a term of
order $(\hat{e} - e)^2$ that the other nuisances cannot offset
(Appendix~\ref{app:verifications}).

The log-scale variant follows from
$\log \tau(x) = \logit(p(x)) - \logit(e(x))$ via the delta method:
\begin{equation}
\label{eq:dr-q-log}
\Gamma^{\text{DR-Q}}_{\log} = \underbrace{\logit(\hat{p}(X)) - \logit(\hat{e}(X))}_{\text{plug-in}} + \underbrace{\frac{Y(W-\hat{p}(X))}{\hat{m}(X) \hat{p}(X)(1-\hat{p}(X))}}_{\text{correction for } p} - \underbrace{\frac{W-\hat{e}(X)}{\hat{e}(X)(1-\hat{e}(X))}}_{\text{correction for } e}
\end{equation}
with the same conditional robustness structure.

\subsection{DR-Q-Simple}
\label{sec:dr-q-simple}

With the propensity known, the $e$-correction of DR-Q has conditional
expectation zero and can be dropped. Analogous to the Q-Simple-Learner
(Section~\ref{sec:q-simple}), this motivates a simplified
construction that operates only on converters ($Y = 1$), eliminates
the need to estimate $m(x)$, and requires only the single nuisance
$\hat{p}(x)$:
\begin{equation}
\label{eq:dr-q-simple}
\Gamma^{\text{DR-Q-Simple}} = \hat{\tau}(X) +
\frac{(1-e(X)) \cdot (W-\hat{p}(X))}{e(X) \cdot (1-\hat{p}(X))^2},
\qquad
\Gamma^{\text{DR-Q-Simple}}_{\log} = \logit(\hat{p}(X)) -
\logit(e(X)) +
\frac{W-\hat{p}(X)}{\hat{p}(X)(1-\hat{p}(X))},
\end{equation}
computed on $Y=1$ observations with the known propensity $e(x)$. Given
exact $e(x)$, both satisfy
$\mathbb{E}[\Gamma \mid X{=}x, Y{=}1] = \tau(x) + O\big((\hat{p}(x) - p(x))^2\big)$
(direct scale; analogously for $\log\tau$ on the log scale). On the
direct scale the bias is exactly
$-\frac{1-e}{e}\cdot\frac{(p-\hat{p})^2}{(1-\hat{p})^2(1-p)} \le 0$, i.e.\
systematically downward. Unlike
DR-S/T and DR-Q, its correction divides by no estimated outcome
probability: only $\hat{p}$ and the known $e$ enter. The
simplification comes at a cost in asymptotic efficiency.
 
\providecommand{\figorpending}[2]{\IfFileExists{figures/#1}{\includegraphics[width=#2]{figures/#1}}{\pending{figure \detokenize{#1}: run Benchmark.ipynb}}}

\section{Empirical Evaluation}
\label{sec:experiments-v2}

We now compare the learners empirically and ask three questions: which
learner estimates a ratio CATE best, both in ranking and in scale; how it
should be tuned; and what limits the estimate when converters are rare. We
study them first on synthetic data, where $\tau(x)$ is known and the
conversion rate and the number of converters can be varied independently,
both with a randomized and with a confounded treatment assignment
(Section~\ref{sec:syn-v2}). We then check the findings on seven randomized
datasets (Section~\ref{sec:rct-v2}). We also ran the benchmark on four
observational datasets, but they hold too little data for the heterogeneity
they carry and allow hardly any conclusion; for brevity, we report them in
Appendix~\ref{app:obs}.
Section~\ref{sec:setup-v2} describes
the benchmark protocol, and Section~\ref{sec:guidance-v2} concludes with
practical recommendations. Details of the benchmark are in
Appendix~\ref{app:details}, further results in Appendices~\ref{app:syn-extra}
and~\ref{app:real-tables}.

\subsection{Setup}
\label{sec:setup-v2}

\subsubsection{Datasets}
\label{sec:datasets-v2}

\paragraph{Synthetic.}
Covariates are bootstrapped from Criteo records. On top of them we place a
baseline $\mu_0(x)$ and a fixed, non-additive effect surface
$\tau(x)$ (median $1.19$, $98\%$ of units in $[0.33, 5.4]$), with known propensity $e = 0.5$
(full specification in Appendix~\ref{app:dgp}). A conversion-rate knob
rescales $\mu_0$ and leaves $\tau(x)$ unchanged, so the conversion rate ($0.2$,
$1$, $5$, $20\%$) and the number of training converters ($50$ to $12{,}800$)
can be varied independently. The training size follows as
$n_{\text{train}} = n_{\text{conv}}/\mathrm{CVR}$, which gives $36$
synthetic datasets along which the behavior of each learner can be traced
($33$ for the tuned variants, whose tuning at the $0.2\%$ rate stops at
$1600$ converters). In a confounded variant, the propensity depends on the
same feature that drives the baseline, $\operatorname{logit} e(x) = 0.8\,
z_0(x)$, clipped to $[0.1, 0.9]$, so that treated units convert more often
irrespective of the effect. The learners do not receive this propensity and
must estimate it (Appendix~\ref{app:dgp}); we run this variant at a conversion rate of $2\%$ and
$800$, $3200$ and $12{,}800$ training converters.

\paragraph{Real-world.}
We select seven randomized datasets, where the propensity is known by design.
They cover a wide range of sample sizes ($n = 64$k--$14.0$M), conversion rates
($0.9\%$--$62\%$) and treatment fractions ($50\%$--$85\%$;
Table~\ref{tab:rct-v2}). To make tuning computationally feasible, we cap the training data at $12{,}800$ converters
where a dataset has more, drawn at its natural conversion rate
(Appendix~\ref{app:protocol}).

\begin{table}[H]
\centering
\caption{Randomized datasets, sorted by conversion rate. Propensity known by
design; converters in the $80\%$ training pool. H(Conv), H(Vis): Hillstrom
with conversion or visit as outcome, both email arms count as treated;
Criteo: outcome is a visit. $^\dagger$Both potential outcomes
observed, treatment simulated with $e = 0.5$.}
\label{tab:rct-v2}
\setlength{\tabcolsep}{6pt}
\begin{tabular}{lrrrrl}
\toprule
Dataset & $n$ & CVR & Treat. & Converters & Domain \\
\midrule
H(Conv) \citep{hillstrom2008} & 64k & 0.9\% & 66.7\% & 462 & Email \\
Twins$^\dagger$ \citep{louizos2017} & 71k & 3.6\% & 50.0\% & 2{,}005 & Neonatal \\
Criteo \citep{diemert2018} & 14.0M & 4.7\% & 85.0\% & 525{,}543 & E-commerce \\
Lenta \citeyearpar{lenta2021} & 687k & 10.8\% & 75.1\% & 59{,}494 & Grocery \\
H(Vis) \citep{hillstrom2008} & 64k & 14.7\% & 66.7\% & 7{,}515 & Email \\
MegaFon \citeyearpar{megafon2019} & 600k & 20.4\% & 50.1\% & 98{,}011 & Telecom \\
X5 Retail \citeyearpar{x5retail2020} & 200k & 62.0\% & 50.0\% & 99{,}202 & Retail \\
\bottomrule
\end{tabular}
\end{table}

\subsubsection{Metrics}
\label{sec:metrics-v2}
We use two metrics, one for each operational use of $\hat\tau(x)$: ranking
for targeting, and calibration for quantifying impact.

\paragraph{Ratio Qini.}
For targeting, only the ranking of the units matters. We measure it by the
Qini coefficient~\citep{radcliffe2011}, adapted to ratio effects by
integrating the ratio rather than the difference of the arm rates:
\[
  \mathrm{Qini} = \int_0^1 \frac{\bar Y^{(1)}(s)}{\bar Y^{(0)}(s)}\,
  \mathrm{d}s - \frac{\bar Y^{(1)}(1)}{\bar Y^{(0)}(1)},
\]
where $\bar Y^{(w)}(s)$ is the observed conversion rate in arm $w$ among the
top-$s$ fraction of the test set ranked by $\hat\tau$, with ties broken at
random; higher is better. On synthetic data we replace the observed rates by the known ones, the means
of $\mu_0(x)$ and $\tau(x)\mu_0(x)$ over the top-$s$ fraction, which removes
the outcome noise, and evaluate on $100{,}000$ test units per seed. On the
observational datasets (Appendix~\ref{app:obs}) we use an inverse-propensity-weighted (IPW) version, which
weights each arm by its inverse estimated propensity. To avoid a division
by zero at the very top of the ranking, the ratio counts as one before the
first control converter, and as the treated rate before the first control
unit.

\paragraph{Calibration.}
For quantifying impact, the scale of $\hat\tau$ matters as well. We measure
calibration on the log scale, where over- and under-prediction count alike
($\log 0.8 = -\log 1.25$), and exponentiate to a multiplicative error factor.
Where $\tau$ is known (synthetic data), we use the pointwise error
\begin{equation}
\label{eq:cal-point-v2}
  \mathrm{cal}_{\text{point}} = \exp\bigl(\mathbb{E}\,|\Delta(x)|\bigr),
  \qquad \Delta(x) = \log\hat\tau(x) - \log\tau(x).
\end{equation}
A value of $1$ is perfect, and $1.2$ means $\hat\tau$ is off by about $20\%$
for a typical unit. Since it compares unit by unit, it also punishes a
wrong ordering. For reference, always predicting the median of $\tau$
gives $1.42$ on our synthetic data. On real data $\tau(x)$ is unknown, so we compare the mean
prediction $\bar\tau_b$ in each of up to $10$ quantile buckets of $\hat\tau$
(ties broken at random, as for the Qini),
sized to hold about $30$ control converters of the test set each, with the
bucket's observed rate ratio $r_b = \bar Y^{(1)}_b / \bar Y^{(0)}_b$:
\begin{equation}
\label{eq:cal-binned-v2}
  \mathrm{cal}_{\text{binned}} = \exp\Bigl(\textstyle\sum_b w_b\,
  \bigl|\log\bar\tau_b - \log r_b\bigr|\Bigr),
\end{equation}
with $w_b$ the bucket share. On synthetic data, where both can be computed,
the binned error orders the learners qualitatively like the pointwise one,
but the absolute values of the two cannot be compared
(Appendix~\ref{app:syn-extra}).

\paragraph{Bias and slope.}
We gain further insight from two diagnostics that describe how the
calibration error arises. The \emph{bias}, the mean of $\Delta(x)$ (binned: of
$\log\bar\tau_b - \log r_b$), measures how far the level of $\hat\tau$ is off;
positive means over-prediction. The \emph{calibration slope}, the OLS
coefficient of $\log\tau$ (binned: $\log r_b$) on $\log\hat\tau$, measures
whether the spread is right: at one it is, below one $\hat\tau$ is too
dispersed, above one it is compressed toward a constant. Two learners with the
same error can thus fail in opposite ways. The same holds across runs, so an
average slope can be close to one although no single run is. We therefore
count the spread as right when the slope lies in the corridor $[0.8, 1.25]$,
within a factor of $1.25$ of one in either direction, and report the share of
runs in which it does. The corridor is symmetric on the log scale, like the
calibration error, and its width is a convention; with $[0.7, 1.4]$ all
shares rise, but the comparisons between the learners stay the same.

\paragraph{Comparing learners.}
All learners share the same train/test split per seed, so we compare them by
paired differences within (dataset, seed). Significance follows from a paired
$t$-test, on real data corrected for the units that the seeds
share~\citep{nadeau2003inference} (Appendix~\ref{app:stest}). With this
many comparisons, a few differences will look significant by chance
alone.

\subsubsection{Learners}
\label{sec:learners-v2}
The benchmark covers four groups. \emph{Plug-in baselines}: the S- and
T-Learners~\citep{kunzel2019metalearners}, with
$\hat\tau = \hat\mu_1/\hat\mu_0$. \emph{Ratio-native learners}: the Q- and
Q-Simple-Learners of Section~\ref{sec:q-learner} and the Neyman-orthogonal
augmentations DR-S, DR-T, DR-Q and DR-Q-Simple of Section~\ref{sec:dr}, each
in direct- and log-scale form. \emph{Difference-CATE comparators}: the X-,
R- and AIPW-DR-Learners, converted via
$\hat\tau = 1 + \hat\tau_{\text{diff}}/\hat\mu_0$. \emph{Tuned learners}: the
S- and Q-Simple-Learners, each tuned on log-loss and on Qini (S-Qini,
Q-Simple-Qini), and the T-Learner, tuned on log-loss
(Section~\ref{sec:tuning-setup-v2}). Every learner that uses a propensity
receives the design propensity where there is one. Only the Q-Learner comes in
both versions: the Q-Simple-Learner takes the design propensity, the Q-Learner
estimates it. This gives $20$ learners on randomized data, synthetic and real.
Where the propensity is unknown (confounded synthetic and observational data),
Q-Simple and both DR-Q-Simple variants drop out ($17$ learners).

\subsubsection{Implementation and tuning}
\label{sec:tuning-setup-v2}
All models are LightGBM~\citep{lightgbm} classifiers or regressors with
library defaults, except for the tuned variants; the R- and all DR-Learners
cross-fit their nuisance models over $5$ folds~\citep{chernozhukov2018double}. Following the defaults of
EconML~\citep{econml}, we apply no truncation of $\hat e$ and no
winsorization of pseudo-outcomes, only numerical guards
(Appendix~\ref{app:tuning}). Each synthetic dataset is run with $10$ seeds,
each real-world dataset with $30$ due to the noisier metrics; results are
averaged over seeds (geometric mean for the calibration error).

Tuning a single outcome model on its log-loss is standard machine learning
practice, so the tuned S-Learner is the benchmark that any meta-learner must
beat to justify its use. For a ratio CATE, however, it has two problems: its
log-loss rewards predicting conversion rather than the effect, and in that
prediction the treatment is only one weak feature among many. We address each
separately: the first by tuning the S-Learner on the ratio Qini of its
$\hat\tau$ instead (S-Qini), as uplift models are commonly selected in
practice, and the second by the T-Learner, which fits one model per arm, tuned
on its log-loss. Finally, we tune our proposal, the Q-Simple-Learner, on its
log-loss and, as a control, on Qini; where the propensity is unknown, the
Q-Learner takes its place. We do not tune the remaining learners, which
combine several nuisance models without a common objective.

Each learner is tuned per dataset by a $100$-trial Optuna~\citep{optuna2019}
search with $5$-fold cross-validation on two additional draws of the data
(Appendix~\ref{app:tuning}); the Qini is IPW-weighted where the propensity is
unknown. We tune the T-Learner only on its log-loss, since Qini tuning would
require a joint search over both arm models.

We do not tune on calibration. In practice only the binned error is available, and
it cannot detect missing heterogeneity: for a constant prediction at the
average ratio the buckets are random subsets of the units, so every bucket's
observed ratio estimates the same average and the constant meets it up to
noise. On our synthetic data it therefore scores a binned error about as low
as the true $\tau(x)$, although its pointwise error is about $1.4$, and lower than
a prediction that ranks correctly but has only half the true spread. As an
objective, the binned error would favor models without heterogeneity;
together with the slope, it serves as a check of a chosen model, not as a
criterion for choosing one. Only for the Q-Simple-Learner does a standard,
optimizable objective measure the error of $\log\hat\tau$ itself: its
log-loss is a converter-weighted squared error of $\log\hat\tau$
(Section~\ref{sec:q-simple}).

\subsection{Synthetic Benchmark}
\label{sec:syn-v2}

\subsubsection{All learners at their defaults}
\label{sec:syn-untuned-v2}
\begin{figure}[H]
\centering
\figorpending{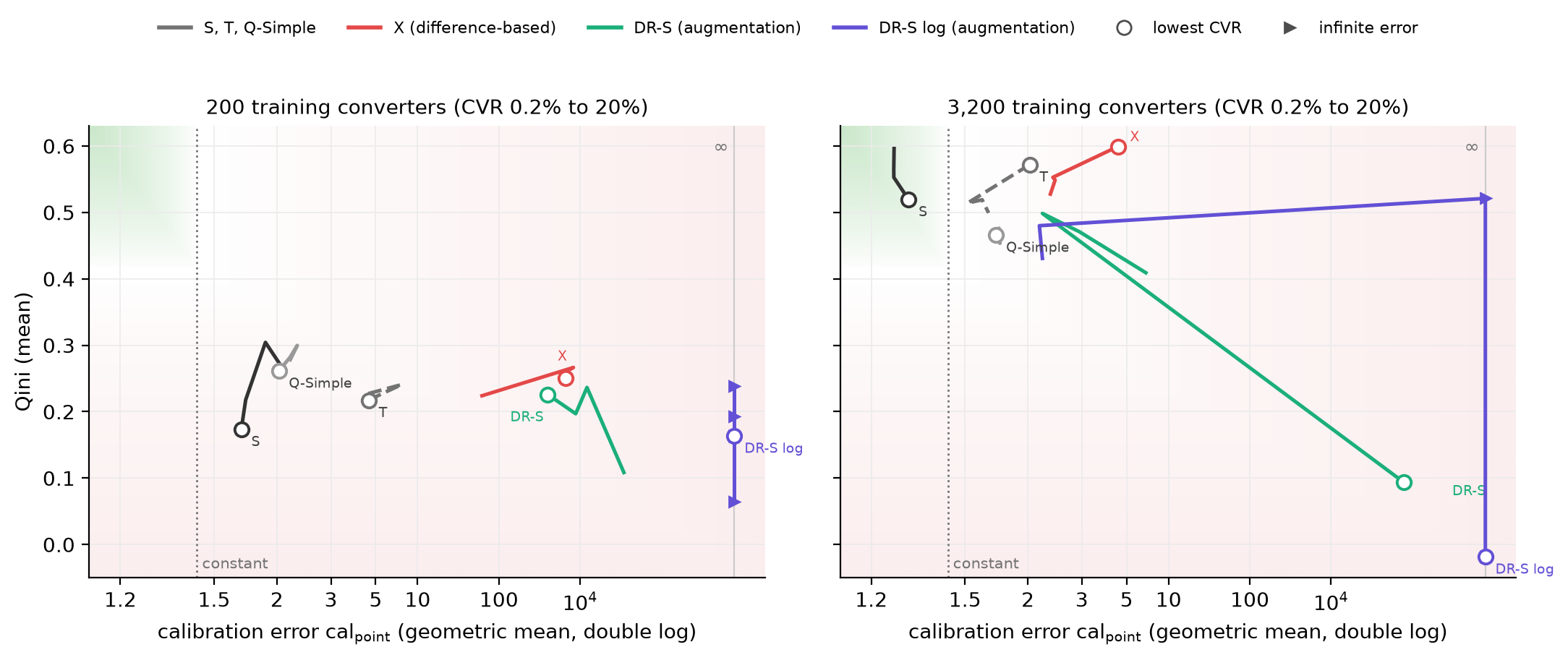}{\textwidth}
\caption{Qini and pointwise calibration error of the untuned learners at
$200$ (left) and $3200$ training converters (right). Each learner forms a line
that connects its values at the different conversion rates. Dotted: constant
predictor at the median of $\tau$; learners to its right are worse calibrated
than a constant. The oracle reaches a Qini of $0.83$. Shown are S, T and
Q-Simple with one representative of each other group (all learners in
Figure~\ref{fig:syn-untuned-all}). \emph{Observation:} the difference-based
X-Learner (red) loses calibration, the augmentations DR-S (green) and DR-S
log (blue) also ranking; from the left to the right panel, the learners
improve far more than along their lines, so the number of converters matters
more than the conversion rate.}
\label{fig:syn-untuned-v2}
\end{figure}

We first run every learner with library defaults.
Figure~\ref{fig:syn-untuned-v2} compares their ranking and calibration at
$200$ and at $3200$ training converters. Each learner is a line connecting
its values at the different conversion rates, so a learner that does not
depend on the rate collapses to a point.

\paragraph{Difference-based learners fall behind the S-, T- and Q-Learners
at calibration, the augmentations also at ranking.}
We attribute the miscalibration of both groups to a division by a small
$\hat\mu$ that, unlike in the ratio $\hat\mu_1/\hat\mu_0$ of the S- and
T-Learners, is not offset by a numerator of the same order: for the
difference-based X-, R- and AIPW-DR-Learners in the conversion
$1 + \hat\tau_{\text{diff}}/\hat\mu_0$, for DR-S, DR-T and DR-Q in the
inverse outcome probabilities of their augmentation terms. It grows the fewer
the converters and the lower the rate. The difference-based learners still
rank about as well as the S- and T-Learners; the augmentations do not: we
find their pseudo-outcomes so heavy-tailed that the second-stage regression
turns them into non-positive (direct scale) or exploding (log scale)
predictions for whole regions of $x$ (Appendix~\ref{app:syn-extra}). Only
DR-Q-Simple on the log scale, which like the Q-Learner never divides by an
estimated outcome probability, stays close to the Q-Simple-Learner in
calibration, though it ranks worse. For a ratio CATE we therefore advise
against both groups.

\paragraph{The number of converters, not the conversion rate, limits how well
a ratio CATE can be learned.}
The Q-Simple-Learner estimates the ratio CATE from the converters alone, up
to the noise of the realized treatment assignment, which becomes negligible
in larger samples. We therefore expect the number of converters to limit the
estimation of a ratio CATE for every learner, as for rare binary events in
general, where the information grows with the number of events rather than
the number of units~\citep{king2001rare,owen2007infinitely}. Our synthetic
benchmark reflects this: although most learners divide by an estimated
baseline, which shrinks with the rate, their ranking and calibration depend
much less on the conversion rate than on the number of converters. In
Figure~\ref{fig:syn-untuned-v2}, each learner's line across the conversion
rates is much shorter than the distance it moves from $200$ to $3200$
converters. Consequently, below we pool the S-, T- and Q-Simple-Learners over
conversion rates at a given number of converters.

\paragraph{The Q- and Q-Simple-Learners coincide.}
With regard to the Qini, the two are identical for the practitioner
(significant differences on $5$ of the $36$ datasets, never more than
$0.08$). Only their calibration differs below about $800$
converters, where the estimated propensity costs the Q-Learner up to $0.38$
log units; from $800$ converters on, the difference is at most $0.012$.
The asymptotic advantage of
Section~\ref{sec:q-simple} thus does not show in our samples, and on
randomized data the Q-Simple-Learner is enough.

\subsubsection{Tuning}
\label{sec:syn-tuning-v2}
\begin{figure}[H]
\centering
\figorpending{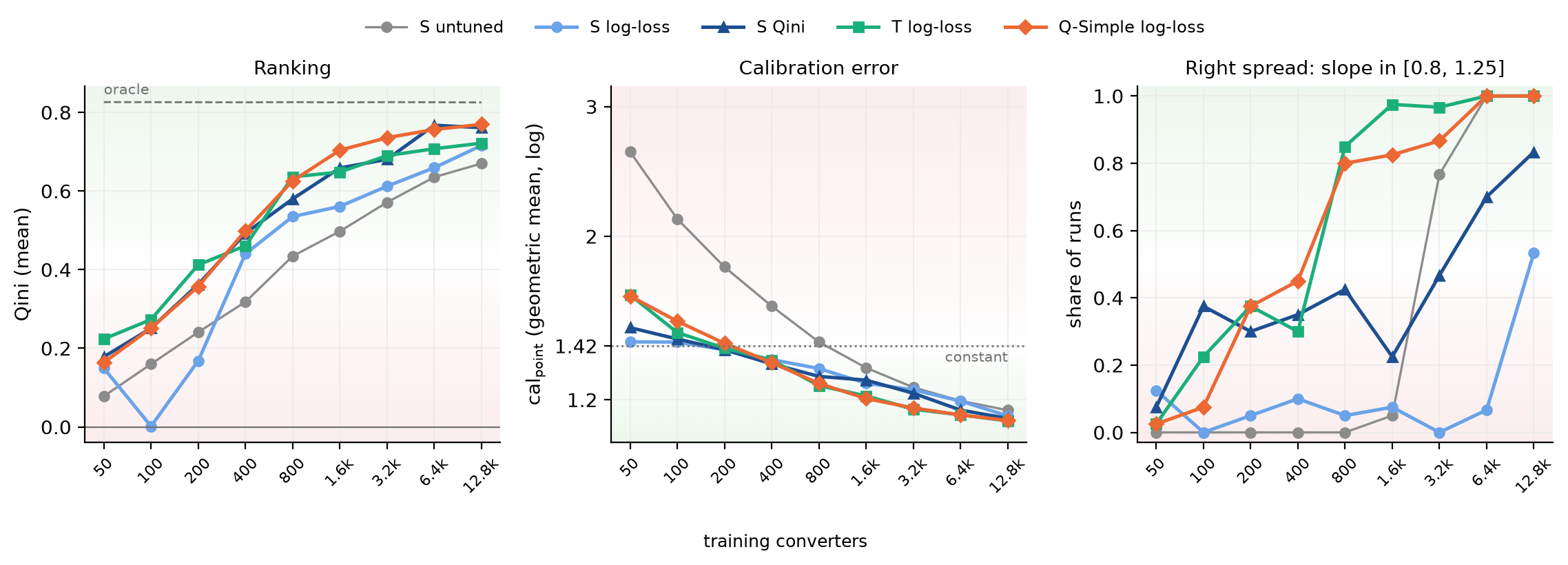}{\textwidth}
\caption{Qini, pointwise calibration error and share of runs with the right
spread (calibration slope in $[0.8, 1.25]$) of the tuned S-, T- and
Q-Simple-Learners and the untuned S-Learner, against the number of training converters, pooled over
conversion rates (from $3200$ converters on without the $0.2\%$ rate). Every
dataset is tuned once, so jumps in the tuned lines reflect different tuning
outcomes rather than seed noise. \emph{Observation:} tuned on log-loss, the
S-Learner falls behind in ranking and spread; tuned on Qini, or replaced by the
log-loss-tuned T-Learner, it catches up in ranking, but only the T- and
Q-Simple-Learners tuned on log-loss get the spread right from about $800$
converters on.}
\label{fig:syn-tuning-main-v2}
\end{figure}

At their defaults, the S-, T- and Q-Learners reach more than half of the
oracle Qini from a few thousand converters on, but they are poorly
calibrated: only the S-Learner beats the constant predictor, from about
$1600$ converters on, and the T- and Q-Learners do so only at the largest
sizes. At its defaults, the Q-Simple-Learner is thus no option; it has to be
tuned. We regard tuning as necessary for any learner and therefore compare
only tuned variants, next to the untuned S-Learner as the reference, from
here on; every
learner is reported in Appendices~\ref{app:syn-extra}
and~\ref{app:real-tables}. Figure~\ref{fig:syn-tuning-main-v2} shows the tuned variants of
Section~\ref{sec:tuning-setup-v2} next to the untuned S-Learner, pooled over
conversion rates.

\paragraph{Tuned on log-loss, the S-Learner falls behind; both remedies work.}
Tuned on its log-loss, the S-Learner settles on an almost constant prediction
when converters are few and ranks poorly or not at all; with more
converters, its spread is rarely right below $12{,}800$. Both remedies of
Section~\ref{sec:tuning-setup-v2} repair the ranking: tuned on Qini, the
S-Learner ranks about as well as the T- and Q-Simple-Learners, and so does the
T-Learner, which no longer has to find the treatment among many features. The
calibration errors of all tuned variants lie close together.

\paragraph{Tuning must be aligned with the ratio CATE; to get the spread
right, it must also be a log-loss.}
Tuned on Qini or, for the T- and Q-Simple-Learners, on their log-loss, all
variants are aligned with the ratio CATE and rank about alike. Only the two
log-loss-tuned variants, T and Q-Simple, also get the spread right from about
$800$ converters on: a log-loss
rewards calibrated probabilities. The Qini, in contrast, is a noisy,
scale-blind criterion that selects among models which rank alike but scale
differently.

\subsubsection{Confounded assignment}
\label{sec:syn-conf-v2}
\begin{figure}[H]
\centering
\figorpending{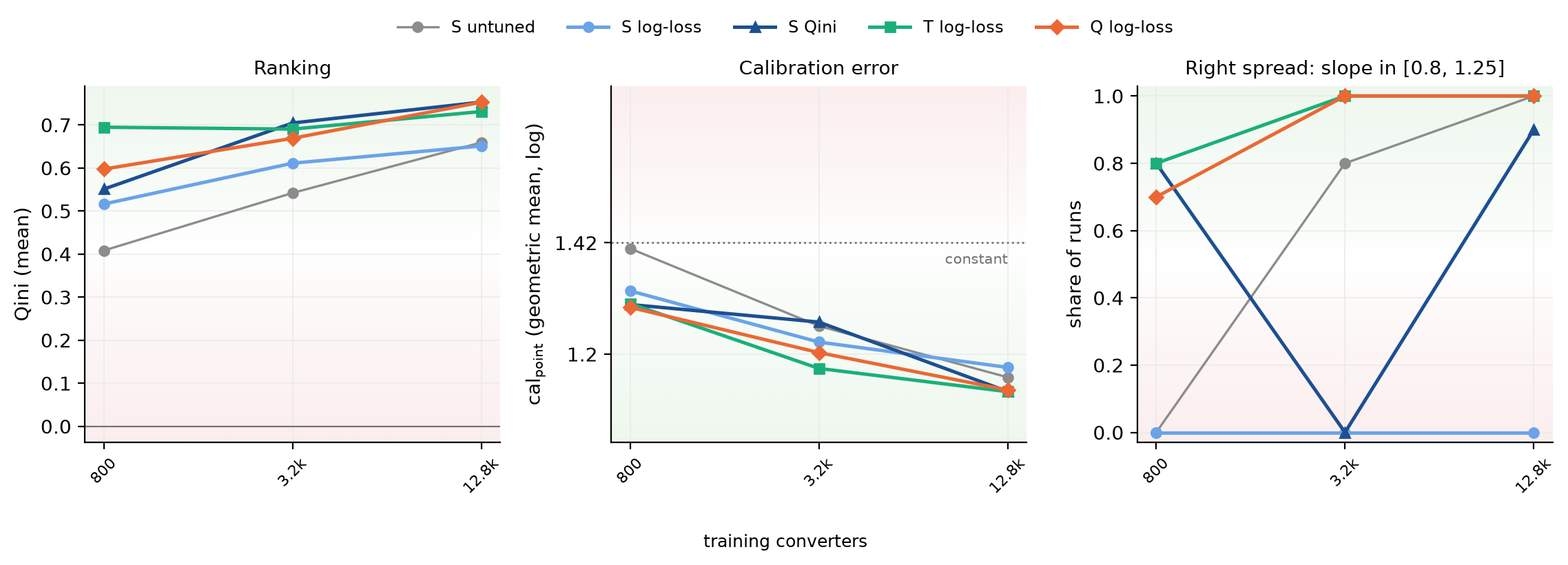}{\textwidth}
\caption{As Figure~\ref{fig:syn-tuning-main-v2}, on the confounded synthetic
data (conversion rate $2\%$, propensity estimated), with the Q-Learner in
place of the Q-Simple-Learner. \emph{Observation:} the picture of the
randomized data repeats; the log-loss-tuned T- and Q-Learners get the spread
right from $3200$ converters on in every run.}
\label{fig:syn-conf-v2}
\end{figure}

We repeat the tuning comparison on the confounded variant
(Section~\ref{sec:datasets-v2}), where the learners must estimate the
propensity, so the Q-Learner takes the Q-Simple-Learner's place. At their
defaults, the learners behave as on the randomized data; every learner is
reported in Tables~\ref{tab:app-synthconf-qini}
and~\ref{tab:app-synthconf-cal}. Figure~\ref{fig:syn-conf-v2} shows the same
variants and metrics as Figure~\ref{fig:syn-tuning-main-v2}.

\paragraph{With an estimated propensity, the picture repeats.}
Tuned on log-loss, the S-Learner never gets the spread right, while all
variants aligned with the ratio CATE rank and calibrate about alike from
$3200$ converters on. Again only the log-loss-tuned T- and Q-Learners get the
spread right reliably, from $3200$ converters on in every run and at $800$ in
most. The estimated
propensity does not hurt the Q-Learner once its propensity model is tuned on
its log-loss. Tuned on Qini instead, the Q-Learner falls behind at $800$
converters in ranking and calibration (Tables~\ref{tab:app-synthconf-qini}
and~\ref{tab:app-synthconf-cal}): the noisy Qini is a poor criterion for
the converter model that enters $\hat\tau$ alongside the propensity model.

\subsection{Randomized Experiments}
\label{sec:rct-v2}
We repeat the analysis on the seven randomized datasets at $12{,}800$
training converters, or the whole pool where it is smaller (H(Conv), Twins
and H(Vis)), to see whether the synthetic findings hold on real effect
structures. Ranking is now measured
by the Qini on observed outcomes and calibration by the binned error, so absolute values
are comparable neither with the synthetic ones nor, for the Qini, between
datasets.

\begin{figure}[H]
\centering
\figorpending{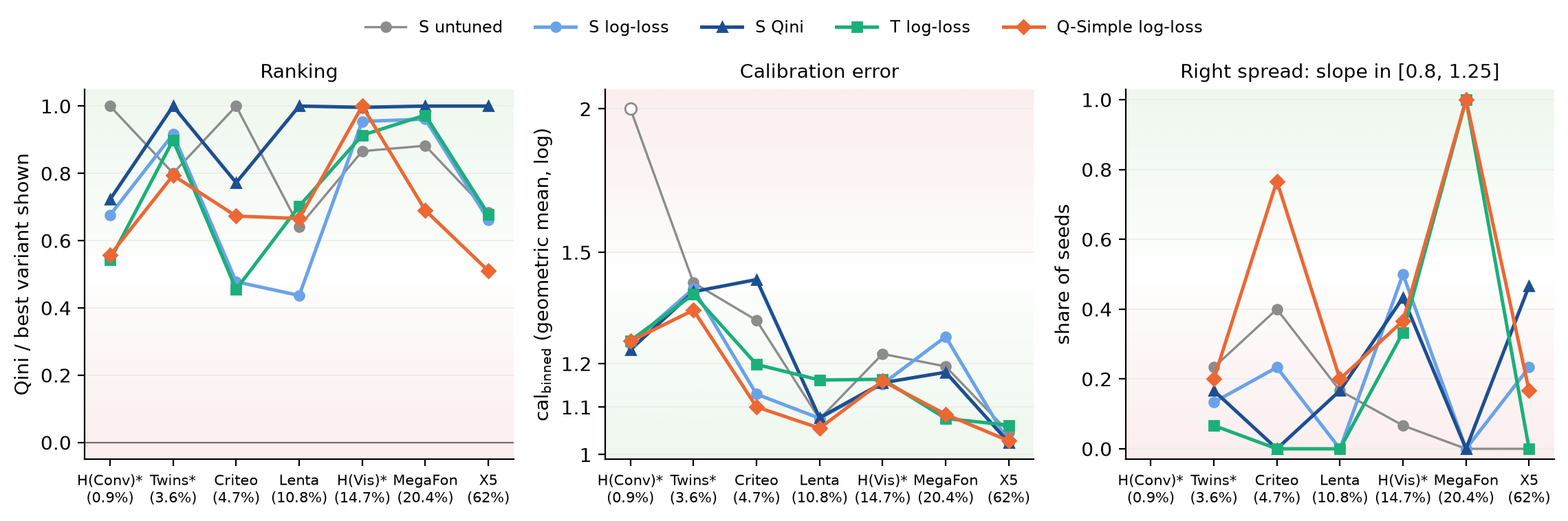}{\textwidth}
\caption{As Figure~\ref{fig:syn-tuning-main-v2}, on the randomized datasets
at $12{,}800$ training converters or the whole pool where it is smaller
(marked $^*$; Table~\ref{tab:rct-v2}), sorted by conversion rate; Qini as a
share of the best variant shown, calibration by the binned error, values
above $2$ pinned at $2$ (open markers). Lines connect the datasets for
readability only. On H(Conv), a single calibration bucket leaves no slope
(Appendix~\ref{app:protocol}). \emph{Observation:} the tuned variants
mostly calibrate alike; where any variant gets the spread right, it is a
Q-Simple variant, on MegaFon also the T-Learner.}
\label{fig:rct-tuning-max-v2}
\end{figure}

At their defaults, the learners behave much as on the synthetic data
(Tables~\ref{tab:app-rct-qini} and~\ref{tab:app-rct-cal}); as there, we
compare the tuned variants next to the untuned S-Learner
(Figure~\ref{fig:rct-tuning-max-v2}).

\paragraph{Again, the Q-Simple-Learner gets the spread right most robustly.}
The calibration errors of the tuned variants mostly lie close together
(exceptions: S-Qini on Criteo and Q-Simple-Qini on H(Conv)), and the
log-loss-tuned Q-Simple-Learner is the best or nearly the best calibrated
variant on every dataset. The spread separates the variants on Criteo and
MegaFon: on MegaFon, both Q-Simple variants and the log-loss-tuned T-Learner
have their slope in the corridor in every seed, the S-Learner variants in none; on
Criteo, only the log-loss-tuned Q-Simple-Learner gets it right, in $77\%$ of
the seeds.
Elsewhere, no variant gets the spread right reliably.

\paragraph{In ranking, tuning helps reliably only for the Q-Simple-Learner.}
Tuned on log-loss, the Q-Simple-Learner ranks significantly better than
untuned on four of the seven datasets and never significantly worse
(Table~\ref{tab:app-rct-qini}). For the S- and T-Learners, tuning is a
gamble: on Criteo, all their tuned variants rank significantly worse than
their untuned learners, and the untuned S-Learner ranks best. Unlike on
synthetic data, the aligned variants do not rank alike: S-Qini ranks best or
close to best on most datasets and significantly better than the
Q-Simple-Learner on MegaFon and X5. The Qini is so noisy a criterion that the
search can pick a model that was merely lucky on the tuning folds.

\subsection{Summary and Practical Guidance}
\label{sec:joint-v2}
\label{sec:guidance-v2}
For a ratio CATE, our benchmark suggests:
\begin{itemize}
  \item \textbf{Always tune, and align the tuning with the ratio CATE.}
  Tuning is necessary to get the scale right; for the ranking, the tuning
  must be aligned with the ratio CATE, as for S-Qini or the log-loss-tuned
  T- and Q(-Simple)-Learners.
  \item \textbf{The Q(-Simple)-Learner is the easiest to use.} On randomized
  data, the Q-Simple-Learner is a single classifier, fitted on the converters
  only and tuned by the standard cross-validated log-loss, at a small
  fraction of the tuning cost of the S- and T-Learners
  (Appendix~\ref{app:tuning}); where the propensity is unknown, the Q-Learner adds a standard propensity model.
  \item \textbf{If the scale matters, use the Q(-Simple)-Learner tuned on its
  log-loss.} From about 200 training converters on, which we regard as the
  standard regime for CATE estimation, it is the best or nearly the best
  calibrated learner on synthetic and randomized data, and where any variant
  gets the spread right reliably, it is among them.
  \item \textbf{If only the ranking matters, there is no clear winner.} On
  synthetic data, all aligned variants rank alike; on randomized real data,
  S-Qini often ranks best, but sometimes the untuned S-Learner ranks better,
  so the candidates need to be compared on one's own data.
\end{itemize}
Our observational datasets are too small to confirm these results
(Appendix~\ref{app:obs}). Whatever the learner, the number of training
converters, not the conversion rate, limits what can be learned: below a few
hundred converters, no learner ranks reliably.

\paragraph{Limitations.}
All learners use LightGBM, and only the S-, T- and Q(-Simple)-Learners are
tuned, the T-Learner only on its log-loss. The other learners may profit from
tuning as well, but for the difference-based learners and the augmentations,
the division by a small $\hat\mu$ remains; tail control such as trimming
$\hat\mu$ might improve them, at the cost of some bias. We compare against no
ratio-native estimator from the literature, such as
contrast regression~\citep{yadlowsky2021ratio}, the EP-Learner~\citep{vanderlaan2024combining},
the R-Learner of \citet{ge2026orthogonal} or the case-only forests of
\citet{dai2019trees}, and we do not evaluate the use of converter-only
features.
On real data, the tuning splits share units with the evaluation test sets,
which may favor the tuned variants, Qini tuning most. Non-positive ratios
from the difference-based learners and the direct-scale augmentations enter
the calibration error at a floor of $10^{-10}$, so for these learners its
magnitude reflects the floor rather
than a typical error. The synthetic data follow a single effect surface,
which with $p = \tau/(1+\tau)$ is a favorable case for the Q-Simple-Learner,
and their rate knob favors the converter-count result by construction; the
confounded variant covers one design at a single conversion rate. Finally,
the observational datasets hold too little data for the heterogeneity they
carry to support conclusions.

\section{Conclusion}
\label{sec:conclusion}

We introduced the Q-Learner, a meta-learner for ratio-based CATEs that
writes $\tau(x)$ as a ratio of two odds via a Bayes-rule
identity, reducing estimation to two binary classification tasks whose
log-loss is aligned with the effect: tuning the classifiers in the standard
way tunes $\hat\tau(x)$ itself. With a known propensity, it reduces to the
Q-Simple-Learner, a single classifier fitted on the converters only, which
applies when detailed data are available only for the converters. We further derived
Neyman-orthogonal augmentations of the Q- and Q-Simple-Learners on the direct
and log scale; since the ratio is a nonlinear functional, their guarantee is
weaker than classical double robustness, as a quadratic remainder survives.

On synthetic and randomized data, our benchmark shows that learners that
divide by an estimated outcome probability without a compensating numerator
lose their calibration
when converters are few, the augmentations without tail control also
their ranking. Tuning proves essential and must be aligned with the ratio
CATE: on synthetic data, the S-Learner tuned as usual on its log-loss falls
behind. Tuned on
its log-loss, the Q(-Simple)-Learner ranks well and gets the scale of the
effects right more reliably than the other learners, at a small fraction of
their tuning cost, while for the ranking alone no learner wins
consistently.

Ultimately, these results point to a single picture: the information about a
ratio effect lives in the converters. In our benchmark, their number, not
the conversion rate, limits what any learner can achieve, and the Q-Learner learns the effect
where it lives, by asking among the converters who was treated. Targeting the
ratio this way, rather than recovering it from outcome models or
differences, turns a fragile division into a standard classification task,
one that modern classifiers solve well and practitioners know how to tune.

Several avenues for future work remain. The most pressing is tail control
for orthogonal ratio learners: our augmentations remove the first-order bias
in theory, but without control of their heavy-tailed pseudo-outcomes they
fail at realistic converter counts, and how to trade this against the bias of
trimming remains open. A second is effect-aligned tuning beyond our three learners: the
Qini-tuned S-Learner and the T-Learner show that aligning the tuning with the
effect, by objective or by structure, lifts learners built for the outcome,
and effect-aligned objectives such as the R-loss may do the same for others. A
third is uncertainty quantification, traditionally hindered by error
propagation in the denominator of ratio-based effects: the influence
functions established here provide a natural foundation for asymptotic
confidence intervals, and for the Q-Learner the delta method on the logit
scale of $\hat p$ and $\hat e$ offers a cheap route to the variance of
$\hat\tau$. Finally, our confounded synthetic data suggest that the Q-Learner copes
with an estimated propensity once its propensity model is tuned;
observational data with enough converters would show whether this carries
over to real confounding.

\paragraph{Code and data availability.}
All code, dataset loaders, and scripts to reproduce the experiments in
this paper are available at
\url{https://github.com/michaelfuchs90/ratiobasedcate}.
The repository includes the result CSVs, the notebook that
generates all figures and tables, and a README with replication
instructions.

\paragraph{Author contributions.}

M.F.\ conceived the project, developed the Q-Learner and the Neyman-orthogonal augmentations, derived the theoretical results, designed and conducted all empirical experiments, and wrote the manuscript. D.K.\ engaged in conceptual discussions and contributed to the structural refinement and finalization of the manuscript.

\paragraph{Acknowledgments.}
We thank Marco Breitig and Kristina Weidner for valuable comments on
the manuscript, and Thomas Nagler, Julian Rodemann, and Lars van der
Laan for detailed feedback on the first arXiv version that shaped
this revision.


\newpage
\appendix

\section{Log-Loss and Calibration of Q-Simple}
\label{app:q-simple}

Throughout, $e$ is known, $\hat\tau=\frac{\hat p}{1-\hat p}\cdot\frac{1-e}{e}$,
$\sigma$ is the logistic function, and
$\Delta(x)=\log\hat\tau(x)-\log\tau(x)=\logit\hat p(x)-\logit p(x)$.
In words: the classifier of the Q-Simple-Learner predicts
$p(x) = P(W{=}1 \mid Y{=}1, X{=}x)$, a fixed monotone function of $\tau(x)$,
and on the logit scale its error is exactly the error $\Delta$ of
$\log\hat\tau$. The log-loss punishes this error roughly quadratically, so a
classifier with a low log-loss has a small error in $\log\hat\tau$
(Proposition~\ref{prop:logloss}). And a classifier whose probabilities are
calibrated, which a low log-loss encourages, gives a calibrated $\hat\tau$
(Proposition~\ref{prop:calibration}).

\begin{proposition}[Log-loss alignment]
\label{prop:logloss}
The excess population log-loss of $\hat p$ among converters is a weighted
squared error of $\log\hat\tau$:
\[
  \mathcal L(\hat p)-\mathcal L(p)
  =\frac{1}{2P(Y=1)}\,\mathbb{E}\big[m(X)\,v(X)\,\Delta(X)^2\big],
\]
with weights $v(x)\in(0,\tfrac14]$ that equal $p(x)(1-p(x))$ up to
$O(|\Delta(x)|)$.
\end{proposition}
\begin{proof}
The excess log-loss is $\mathbb{E}[\mathrm{KL}(p(X)\,\|\,\hat p(X))\mid Y=1]$.
With $A(\eta)=\log(1+e^\eta)$, $\eta=\logit p$, $\hat\eta=\logit\hat p$:
$\mathrm{KL}(p\|\hat p)=A(\hat\eta)-A(\eta)-A'(\eta)(\hat\eta-\eta)=\tfrac12A''(\xi)\Delta^2$
by Taylor's theorem for some $\xi$ between $\eta$ and $\hat\eta$, and
$v=A''(\xi)=\sigma'(\xi)$. The factor $m/P(Y=1)$ follows from
$\mathbb{E}[g(X)\mid Y=1]=\mathbb{E}[g(X)m(X)]/P(Y=1)$.
\end{proof}

\begin{proposition}[Calibration transfer]
\label{prop:calibration}
Let $W\perp X$ with $P(W=1)=e$. If $\hat p$ is calibrated among converters,
$P(W=1\mid\hat p(X)=q,Y=1)=q$ for all $q$, then
\[
  \frac{\mathbb{E}[Y\mid W=1,\hat\tau(X)=t]}{\mathbb{E}[Y\mid W=0,\hat\tau(X)=t]}=t\quad\text{for all }t .
\]
\end{proposition}
\begin{proof}
Let $S=\{\hat p(X)=q\}=\{\hat\tau(X)=t\}$. By calibration and $W\perp X$,
$\frac{q}{1-q}=\frac{P(W=1,Y=1\mid S)}{P(W=0,Y=1\mid S)}=\frac{e\,\mathbb{E}[\mu_1(X)\mid S]}{(1-e)\,\mathbb{E}[\mu_0(X)\mid S]}$,
and $\mathbb{E}[Y\mid W=w,S]=\mathbb{E}[\mu_w(X)\mid S]$.
\end{proof}

Minimizing the converter log-loss thus minimizes a weighted squared error
of $\log\hat\tau$, with weights proportional to $m(x)\,p(x)(1-p(x))$: the fit
is the effect error, but measured where the converters are. Regions with a
low conversion probability carry little weight, which is the price of
learning from converters only. For an outcome model,
$\log\hat\tau-\log\tau$ is the difference of two arm-wise log errors, and
its log-loss is dominated by the baseline fit that both arms share. For
the two-model Q-Learner, $\Delta=\Delta_p-\Delta_e$ with
$\Delta_p=\logit\hat p-\logit p$ and $\Delta_e=\logit\hat e-\logit e$, and
the two log-losses
control $|\Delta|$ only through $|\Delta_p|+|\Delta_e|$.

\section{Robustness Verifications for Section~\ref{sec:dr}}
\label{app:verifications}

With $\hat A=\hat p/(1-\hat p)$, $\hat B=(1-\hat e)/\hat e$ and all
functions evaluated at $x$, direct computation of the conditional
expectations gives
\begin{align}
\E[\Gamma^{\text{DR-S/T}}\mid X{=}x]-\tau
 &=\frac{(\tau-\hat\tau)(\mu_0-\hat\mu_0)}{\hat\mu_0}
  +\frac{e-\hat e}{\hat\mu_0}\Big[\frac{\mu_1-\hat\mu_1}{\hat e}+\frac{\hat\tau(\mu_0-\hat\mu_0)}{1-\hat e}\Big],\\
\E[\Gamma^{\text{DR-Q}}\mid X{=}x]-\tau
 &=\frac{\hat B(p-\hat p)}{(1-\hat p)^2}\Big[\frac{m-\hat m}{\hat m}-\frac{p-\hat p}{1-p}\Big]
  \nonumber\\
 &\qquad+\frac{(p-\hat p)(e-\hat e)}{(1-p)(1-\hat p)\,e\,\hat e}
  -\frac{\hat A(e-\hat e)^2}{e\,\hat e^2},\\
\E[\Gamma^{\text{DR-Q-Simple}}\mid X{=}x,Y{=}1]-\tau
 &=-\frac{1-e}{e}\cdot\frac{(p-\hat p)^2}{(1-\hat p)^2(1-p)}\;\le 0 .
\end{align}
Every term is a product of two nuisance errors. DR-S/T is conditionally
unbiased if the outcome models are correct, whatever $\hat e$; if only
$\hat e$ is correct, the bias $(\tau-\hat\tau)(\mu_0-\hat\mu_0)/\hat\mu_0$
remains, so it is not doubly robust in the classical sense. DR-Q is
conditionally unbiased if $\hat p$ and $\hat e$ are correct, whatever
$\hat m$; a misspecified $\hat e$ leaves the term
$-\hat A(e-\hat e)^2/(e\hat e^2)$, which is second order but cannot be
offset by the other nuisances. The log-scale variants have the same
structure.

\section{Benchmark Details}
\label{app:details}

\subsection{Synthetic data-generating process}
\label{app:dgp}
Every seed draws a fresh sample of $n_{\text{train}}$ training and
$100{,}000$ test units. Each
unit is a row drawn with replacement from a fixed random subsample of
$500{,}000$ Criteo records, restricted to eight of its twelve features
($f_0, f_2, f_3, f_5, f_6, f_7, f_9, f_{10}$), dropping two that are nearly
constant and two that are nearly redundant with retained ones; the learners see these eight
raw features. For the outcome model we standardize them over the subsample to
$z_k$, with mean zero and unit variance, and set
\begin{align*}
  \mu_0(x) &= s\,\sigma\bigl(b(x)\bigr), &
  b(x) &= S_{0,\,1.2}\bigl(z_0 - 0.8\,z_2 + 0.6\,z_6\bigr),\\
  \mu_1(x) &= \tau(x)\,\mu_0(x), &
  \log\tau(x) &= S_{0.25,\,0.5}\bigl(0.5\,z_2 - 0.4\,z_6 + 0.4\,z_2 z_9
                 + 0.3\tanh z_0\bigr),
\end{align*}
where $\sigma$ is the logistic function and $S_{m,\,v}$ rescales a surface
over the drawn sample to mean $m$ and standard deviation $v$. The effect
thus uses a different feature mix than the baseline, includes an interaction
and a nonlinearity, and is held fixed across the grid. The scale $s$ is solved
for numerically so that the marginal conversion rate
$\tfrac12\,\mathbb{E}[\mu_1] + \tfrac12\,\mathbb{E}[\mu_0]$ matches the target
rate; this is the conversion-rate knob, which leaves $\tau(x)$ unchanged.
Both $\mu_0$ and $\mu_1$ are bounded to $[10^{-5}, 1 - 10^{-5}]$. Treatment is
assigned with $e = 0.5$ and outcomes are drawn as
$Y \sim \mathrm{Bernoulli}(\mu_W(x))$. On the test units, the population metrics use $\mu_0(x)$ and $\tau(x)$ directly.
\paragraph{Confounded variant.} It uses the same baseline and effect but
assigns treatment with $e(x) = \sigma(0.8\, z_0)$, clipped to $[0.1, 0.9]$.
Since $z_0$ raises the baseline, and weakly the effect, treated units
convert more often than controls irrespective of the effect, so that a
learner ignoring the assignment is biased. The scale $s$ now matches
$\mathbb{E}[e(x)\mu_1 + (1-e(x))\mu_0]$ to the target rate of $2\%$. The
learners do not receive $e(x)$ and must estimate it, and the IPW-weighted
Qini is used for tuning; the evaluation uses the population metrics as
above. We run it at $800$, $3200$ and $12{,}800$ training converters with
$10$ seeds each.

\subsection{Protocol on real data}
\label{app:protocol}
Each seed splits the data at random into $80\%$ training pool and $20\%$
test set; on Criteo the test set is capped at $500{,}000$ rows. Where the pool
holds more than $12{,}800$ converters (Criteo, Lenta, MegaFon, X5), the
training set is drawn from it stratified by the outcome: exactly $12{,}800$
converters and as many non-converters as the dataset's conversion rate
implies. Sampling uniformly within each outcome stratum leaves $P(W \mid X, Y)$
unchanged, so the design propensity remains valid for the Q-Simple-Learner.
The cap of $12{,}800$ converters matches the largest synthetic size and keeps
the tuning of the S-Learner, by far the most expensive, feasible. The other
three randomized datasets and all observational ones are trained on the whole
pool. The smallest pool, H(Conv) with $462$ converters, leaves only about $25$
control converters in a test set, so its Qini is close to noise and its
binned calibration has a single bucket.

\subsection{Implementation and tuning}
\label{app:tuning}
The numerical guards are: fitted probabilities are bounded away from zero by
$10^{-10}$, and the R-Learner's treatment residual by $10^{-5}$ in absolute
value. The second stage of the augmentations regresses the pseudo-outcome on
$x$ by least squares, since on the direct scale it can be negative. Their
direct-scale predictions can therefore be negative too, as can the ratio
recovered by the difference-based learners. To still obtain a calibration
error, we floor such values at $10^{-10}$ inside the logarithm of
Equations~\eqref{eq:cal-point-v2} and~\eqref{eq:cal-binned-v2}.
In the Qini and
in the calibration buckets, ties in $\hat\tau$ are broken at random with a
fixed seed, so that a learner with large blocks of identical predictions
cannot score by the order of the data.

Each tuned learner is tuned once per dataset by $100$ trials of Optuna's TPE
sampler (seed $0$) over the eight LightGBM hyperparameters of
Table~\ref{tab:app-search-space}. The first trial is a fixed
configuration\footnote{\texttt{num\_leaves} $7$ (S- and T-Learner) or $2$
(Q-Simple- and Q-Learner), \texttt{min\_child\_samples} $20$,
\texttt{min\_child\_weight} $10^{-3}$, \texttt{learning\_rate} $0.1$,
\texttt{n\_estimators} $100$, \texttt{reg\_lambda} $10^{-3}$, no row or
column subsampling.}, the remaining $99$ are proposed by TPE, and the best
trial is kept.
A trial is scored by $5$-fold cross-validation on each of two tuning draws,
generated exactly like the evaluation data but with seeds disjoint from the
evaluation seeds, and averaged over both. On synthetic data these are fresh
samples; on real data they are other random splits and therefore share units
with the evaluation test sets. The log-loss variants score the
model's own classification task: $Y$ on $(X, W)$ for the S-Learner, $Y$ on $X$ in
each arm for the T-Learner, $W$ on $X$ among the converters for the
Q-Simple-Learner. Where the propensity is unknown (observational and
confounded synthetic data), the Q-Learner's converter model takes the parameters tuned for that task, and its
propensity model ($W$ on $X$, all units) is tuned separately on log-loss. The
Qini variants score the ratio Qini of the learner's $\hat\tau$ on the
held-out fold, with folds stratified by $(W, Y)$; there, the
Qini is IPW-weighted with a propensity model fitted on the training folds, and
the Qini-tuned Q-Learner keeps the log-loss-tuned propensity model and searches
the converter model only.
On the synthetic datasets, tuning all $33$ datasets took about $1.3$ hours for
the Q-Simple-Learner on log-loss, $6$ hours on Qini, $35$ hours for the
S-Learner on log-loss, $47$ hours on Qini, and $58$ hours for the T-Learner,
whose two arm models are tuned separately.

\begin{table}[H]
\centering
\caption{Search space of the tuned LightGBM models ($\log$: sampled on a log
scale). The row subsample is drawn anew in every iteration.}
\label{tab:app-search-space}
\footnotesize
\begin{tabular}{ll}
\toprule
Hyperparameter & Range \\
\midrule
\texttt{num\_leaves} & $2$--$128$ ($\log$) \\
\texttt{min\_child\_samples} & $2$--$50$ \\
\texttt{min\_child\_weight} & $10^{-6}$--$10^{-1}$ ($\log$) \\
\texttt{learning\_rate} & $0.005$--$0.2$ ($\log$) \\
\texttt{n\_estimators} & $50$--$1000$ ($\log$) \\
\texttt{reg\_lambda} & $10^{-3}$--$20$ ($\log$) \\
\texttt{subsample} & $0.5$--$1$ \\
\texttt{colsample\_bytree} & $0.5$--$1$ \\
\bottomrule
\end{tabular}
\end{table}

\subsection{Correlation between seeds}
\label{app:stest}
Two seeds of a real dataset are not independent: their random test sets share
units, and on whole-pool datasets their training sets overlap. For the paired
differences $d_1, \dots, d_J$ of two learners over $J$ seeds, with sample
variance $s^2$ and correlation $\rho$ between seeds, the corrected resampled
$t$-test of \citet{nadeau2003inference} replaces the variance of their mean
by
\[
  \widehat{\mathrm{Var}}(\bar d) = s^2 \Bigl(\frac{1}{J} + \frac{\rho}{1-\rho}\Bigr),
\]
and we call a difference significant if $|\bar d| > 2\,
\widehat{\mathrm{Var}}(\bar d)^{1/2}$. On synthetic data every seed draws a
fresh sample, so $\rho = 0$ and the test is the plain paired $t$-test. On
real data we estimate $\rho$ per dataset, learner pair and metric as
$\rho = o_\text{test}\, s_\text{test} + o_\text{train}(1 - s_\text{test})$.
Here $s_\text{test}$ is the share of the variance of the paired difference
over seeds that is due to the test draw: for five seeds we refit the six S-
and Q-Simple variants exactly as in the benchmark, bootstrap the test set
$200$ times with the predictions held fixed, and divide the mean bootstrap
variance by the variance over all $30$ seeds. $o_\text{test}$ is the chance
that a test unit of one seed is a test unit of another ($0.2$, on Criteo
$0.036$ because of the cap of $500{,}000$ test rows), and $o_\text{train}$ the
same for training converters ($12{,}800$ over the pool's converters).
Treating all non-test variance as shared in proportion to $o_\text{train}$,
and ignoring the finite-population correction of the bootstrap, both make the
estimate conservative. We use the largest estimate over these pairs per
dataset and metric for all learner pairs; datasets trained on the whole pool, and all observational
ones, keep the classic factor $\rho/(1-\rho) = n_\text{test}/n_\text{train} =
0.25$.
The test draw dominates the variance of the paired differences (a median
share of $0.88$ for the Qini and $0.70$ for the log calibration error), so on
real data the number of converters in the evaluation sample, not the number
of seeds, limits what can be resolved.

\section{Additional Synthetic Results}
\label{app:syn-extra}

\paragraph{Is the binned calibration a usable proxy?}
On real data only the binned calibration error of Equation~\eqref{eq:cal-binned-v2}
is available, so we check it against the pointwise version on the synthetic
datasets, where both can be computed. Across all runs the two errors have a
rank correlation of $0.90$, and within a dataset the two orderings of the
$20$ learners, by their mean log error, agree with a median Kendall $\tau$
of $0.85$. For the comparisons the main text relies on (tuned against
untuned, S against Q-Simple, log-loss against Qini tuning, Q against
Q-Simple), the binned error confirms four fifths of the $283$ significant
pointwise differences and misses the rest, i.e.\ it has less power. It
significantly reverses the sign in four cases: three of a tuned against the
untuned S-Learner from $3200$ converters on, where the pointwise differences
are below $0.02$ log units, and one of the S- against the Q-Simple-Learner
at the $0.2\%$ rate. The median binned slope is close to the pointwise one,
and the binned bias lies a median $0.13$ log units from the pointwise bias.
The absolute levels differ: even the true
$\tau(x)$ scores a binned error from $1.04$ at a $20\%$ to $1.25$ at a $0.2\%$
conversion rate, since the test set has a fixed size. We therefore read the
binned error only together with its slope, and only to compare learners.

\begin{figure}[H]
\centering
\figorpending{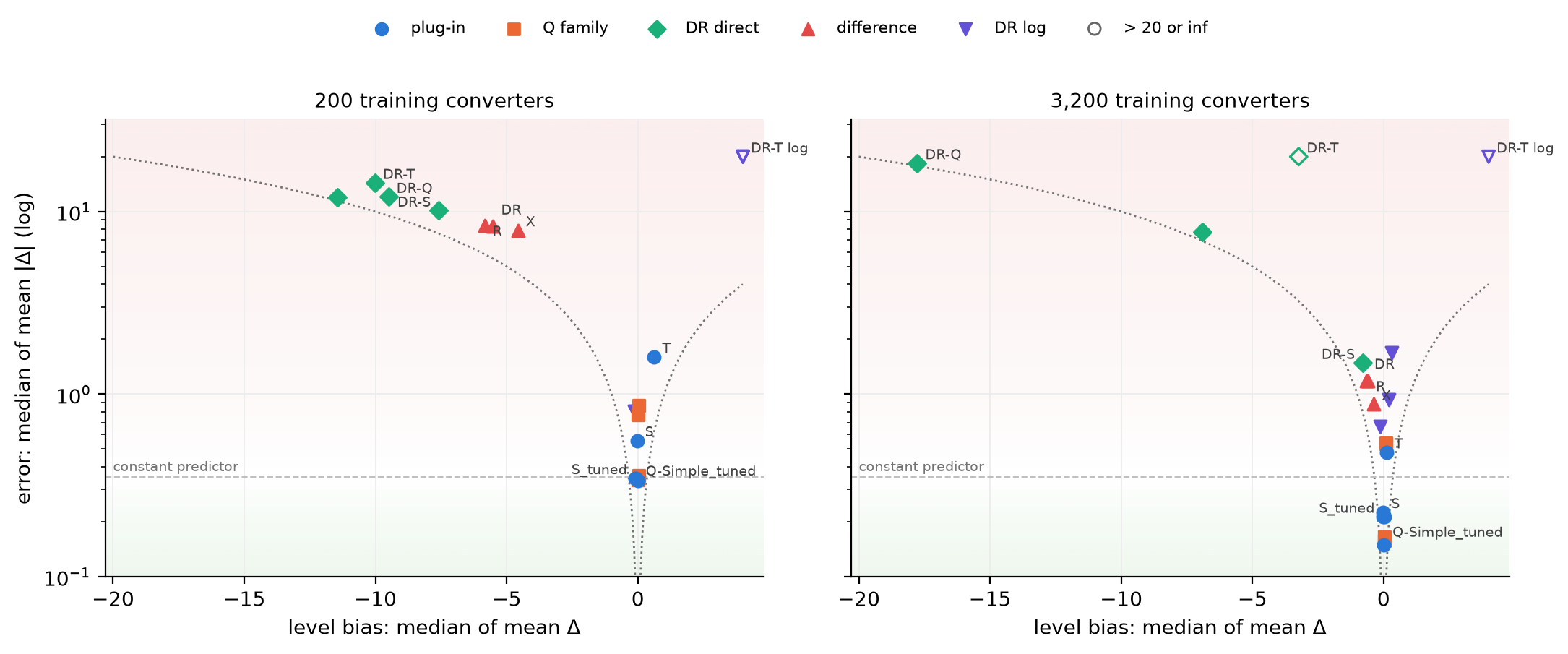}{\textwidth}
\caption{Two ways to lose calibration, pooled over $0.2\%$, $1\%$, $5\%$ and $20\%$;
left $200$, right $3200$ training converters. Each point is a learner; $x$:
median level bias $\mathbb{E}\Delta$; $y$: median mean absolute log error
$\mathbb{E}|\Delta| = \log\mathrm{cal}_\text{point}$ (log axis). Dashed:
constant predictor. Open marker: error above $20$ or infinite.
\emph{Observation:} the difference-based learners sit far above the lines,
an error that differs from unit to unit; most direct-scale augmentations sit
on the left line, an underprediction of almost every unit.}
\label{fig:syn-bias-v2}
\end{figure}

\paragraph{Two ways to lose calibration.}
Figure~\ref{fig:syn-bias-v2} separates the level of the log error from its
size. Because $\mathbb{E}|\Delta| \ge |\mathbb{E}\Delta|$, no learner can lie
below the dotted lines $y = |x|$. A learner on them errs in the same
direction for almost every unit; a learner far above them is wrong unit by
unit in both directions. The difference-based learners lie far above the
lines at small converter counts; for the X-Learner at $1\%$ and $200$
converters, $29\%$ of the units receive a non-positive ratio. DR-T, DR-Q and
DR-Q-Simple on the direct scale lie on the left-hand line, with median level
biases of $-3$ to $-18$ log units even at $3200$ converters: a large share of their predictions is non-positive and
enters the error at the floor of $10^{-10}$. The log-scale DR-S, DR-T and
DR-Q have an infinite error in about two thirds of the runs, since single
predictions explode; only DR-Q-Simple on the log scale stays close to zero
bias.

\paragraph{Every learner along the converter axis.}
Tables~\ref{tab:app-syn-qini} and~\ref{tab:app-syn-cal} report every learner
at every number of training converters, pooled over the conversion rates; how
the rate shifts the learners is shown in Figure~\ref{fig:syn-untuned-v2}.

\begin{table}[H]
\centering
\caption{Population Qini on the synthetic benchmark by number of training
converters, pooled over the conversion rates ($0.2\%$ only up to $1600$
converters; mean over $40$ runs, from $3200$ converters on over $30$; standard
error in parentheses). The oracle reaches about $0.83$. Best per column in
bold; $^\dagger$: tuned variant significantly different from its untuned
learner.}
\label{tab:app-syn-qini}
\footnotesize
\setlength{\tabcolsep}{3pt}
\begin{tabular}{l|rrrrrrrrr}
\hline
Learner & 50 & 100 & 200 & 400 & 800 & 1600 & 3200 & 6400 & 12800 \\
\hline
S & 0.078 & 0.160 & 0.241 & 0.318 & 0.434 & 0.497 & 0.571 & 0.635 & 0.670 \\
 & \scriptsize (0.019) & \scriptsize (0.016) & \scriptsize (0.017) & \scriptsize (0.013) & \scriptsize (0.011) & \scriptsize (0.008) & \scriptsize (0.007) & \scriptsize (0.007) & \scriptsize (0.004) \\[2pt]
T & 0.079 & 0.159 & 0.227 & 0.294 & 0.395 & 0.451 & 0.511 & 0.590 & 0.635 \\
 & \scriptsize (0.018) & \scriptsize (0.014) & \scriptsize (0.012) & \scriptsize (0.010) & \scriptsize (0.008) & \scriptsize (0.008) & \scriptsize (0.006) & \scriptsize (0.006) & \scriptsize (0.004) \\[2pt]
\hline
Q & 0.101 & 0.231 & 0.274 & 0.283 & 0.302 & 0.383 & 0.465 & 0.548 & 0.600 \\
 & \scriptsize (0.020) & \scriptsize (0.019) & \scriptsize (0.017) & \scriptsize (0.010) & \scriptsize (0.010) & \scriptsize (0.008) & \scriptsize (0.007) & \scriptsize (0.006) & \scriptsize (0.005) \\[2pt]
Q-Simple & 0.103 & 0.260 & 0.280 & 0.273 & 0.297 & 0.383 & 0.463 & 0.547 & 0.600 \\
 & \scriptsize (0.028) & \scriptsize (0.024) & \scriptsize (0.019) & \scriptsize (0.011) & \scriptsize (0.011) & \scriptsize (0.009) & \scriptsize (0.008) & \scriptsize (0.006) & \scriptsize (0.006) \\[2pt]
\hline
X & 0.118 & 0.189 & 0.251 & 0.306 & 0.405 & 0.474 & 0.543 & 0.628 & 0.673 \\
 & \scriptsize (0.016) & \scriptsize (0.011) & \scriptsize (0.012) & \scriptsize (0.009) & \scriptsize (0.008) & \scriptsize (0.008) & \scriptsize (0.006) & \scriptsize (0.006) & \scriptsize (0.004) \\[2pt]
R & 0.091 & 0.164 & 0.218 & 0.276 & 0.370 & 0.438 & 0.503 & 0.593 & 0.646 \\
 & \scriptsize (0.012) & \scriptsize (0.011) & \scriptsize (0.012) & \scriptsize (0.008) & \scriptsize (0.007) & \scriptsize (0.010) & \scriptsize (0.007) & \scriptsize (0.007) & \scriptsize (0.005) \\[2pt]
AIPW-DR & 0.096 & 0.168 & 0.221 & 0.279 & 0.372 & 0.440 & 0.502 & 0.592 & 0.648 \\
 & \scriptsize (0.011) & \scriptsize (0.011) & \scriptsize (0.012) & \scriptsize (0.008) & \scriptsize (0.008) & \scriptsize (0.010) & \scriptsize (0.007) & \scriptsize (0.007) & \scriptsize (0.004) \\[2pt]
\hline
DR-S & 0.096 & 0.149 & 0.192 & 0.200 & 0.284 & 0.263 & 0.459 & 0.581 & 0.641 \\
 & \scriptsize (0.014) & \scriptsize (0.013) & \scriptsize (0.014) & \scriptsize (0.014) & \scriptsize (0.016) & \scriptsize (0.028) & \scriptsize (0.015) & \scriptsize (0.010) & \scriptsize (0.009) \\[2pt]
DR-T & 0.066 & 0.123 & 0.162 & 0.163 & 0.200 & 0.154 & 0.172 & 0.168 & 0.161 \\
 & \scriptsize (0.012) & \scriptsize (0.015) & \scriptsize (0.017) & \scriptsize (0.014) & \scriptsize (0.019) & \scriptsize (0.016) & \scriptsize (0.020) & \scriptsize (0.013) & \scriptsize (0.012) \\[2pt]
DR-Q & 0.097 & 0.138 & 0.188 & 0.194 & 0.206 & 0.178 & 0.187 & 0.166 & 0.178 \\
 & \scriptsize (0.014) & \scriptsize (0.012) & \scriptsize (0.014) & \scriptsize (0.016) & \scriptsize (0.017) & \scriptsize (0.013) & \scriptsize (0.019) & \scriptsize (0.010) & \scriptsize (0.016) \\[2pt]
DR-Q-Simple & 0.104 & 0.132 & 0.087 & 0.089 & 0.061 & 0.101 & 0.157 & 0.203 & 0.275 \\
 & \scriptsize (0.022) & \scriptsize (0.032) & \scriptsize (0.023) & \scriptsize (0.013) & \scriptsize (0.009) & \scriptsize (0.010) & \scriptsize (0.010) & \scriptsize (0.011) & \scriptsize (0.013) \\[2pt]
\hline
DR-S log & 0.065 & 0.116 & 0.165 & 0.182 & 0.282 & 0.285 & 0.477 & 0.581 & 0.637 \\
 & \scriptsize (0.016) & \scriptsize (0.012) & \scriptsize (0.016) & \scriptsize (0.015) & \scriptsize (0.014) & \scriptsize (0.027) & \scriptsize (0.011) & \scriptsize (0.010) & \scriptsize (0.008) \\[2pt]
DR-T log & 0.024 & 0.110 & 0.147 & 0.110 & 0.156 & 0.127 & 0.269 & 0.480 & 0.595 \\
 & \scriptsize (0.014) & \scriptsize (0.014) & \scriptsize (0.017) & \scriptsize (0.014) & \scriptsize (0.014) & \scriptsize (0.014) & \scriptsize (0.017) & \scriptsize (0.013) & \scriptsize (0.009) \\[2pt]
DR-Q log & 0.071 & 0.126 & 0.164 & 0.150 & 0.177 & 0.201 & 0.266 & 0.459 & 0.579 \\
 & \scriptsize (0.012) & \scriptsize (0.015) & \scriptsize (0.018) & \scriptsize (0.015) & \scriptsize (0.015) & \scriptsize (0.018) & \scriptsize (0.020) & \scriptsize (0.015) & \scriptsize (0.009) \\[2pt]
DR-Q-Simple log & 0.109 & 0.217 & 0.198 & 0.179 & 0.130 & 0.213 & 0.295 & 0.407 & 0.496 \\
 & \scriptsize (0.025) & \scriptsize (0.026) & \scriptsize (0.017) & \scriptsize (0.012) & \scriptsize (0.008) & \scriptsize (0.010) & \scriptsize (0.009) & \scriptsize (0.007) & \scriptsize (0.006) \\[2pt]
\hline
S (log-loss) & 0.149$^\dagger$ & 0.002$^\dagger$ & 0.168$^\dagger$ & 0.440$^\dagger$ & 0.535$^\dagger$ & 0.561$^\dagger$ & 0.612$^\dagger$ & 0.659$^\dagger$ & 0.716$^\dagger$ \\
 & \scriptsize (0.031) & \scriptsize (0.036) & \scriptsize (0.034) & \scriptsize (0.021) & \scriptsize (0.013) & \scriptsize (0.008) & \scriptsize (0.006) & \scriptsize (0.007) & \scriptsize (0.004) \\[2pt]
S (Qini) & 0.178$^\dagger$ & 0.251$^\dagger$ & 0.363$^\dagger$ & 0.492$^\dagger$ & 0.580$^\dagger$ & 0.659$^\dagger$ & 0.681$^\dagger$ & \textbf{0.767}$^\dagger$ & 0.761$^\dagger$ \\
 & \scriptsize (0.027) & \scriptsize (0.024) & \scriptsize (0.032) & \scriptsize (0.019) & \scriptsize (0.016) & \scriptsize (0.010) & \scriptsize (0.010) & \scriptsize (0.006) & \scriptsize (0.004) \\[2pt]
T (log-loss) & \textbf{0.224}$^\dagger$ & \textbf{0.273}$^\dagger$ & \textbf{0.412}$^\dagger$ & 0.461$^\dagger$ & \textbf{0.636}$^\dagger$ & 0.648$^\dagger$ & 0.690$^\dagger$ & 0.707$^\dagger$ & 0.722$^\dagger$ \\
 & \scriptsize (0.029) & \scriptsize (0.021) & \scriptsize (0.021) & \scriptsize (0.012) & \scriptsize (0.013) & \scriptsize (0.008) & \scriptsize (0.006) & \scriptsize (0.006) & \scriptsize (0.003) \\[2pt]
Q-Simple (log-loss) & 0.164$^\dagger$ & 0.251 & 0.357$^\dagger$ & 0.499$^\dagger$ & 0.626$^\dagger$ & \textbf{0.704}$^\dagger$ & \textbf{0.736}$^\dagger$ & 0.756$^\dagger$ & \textbf{0.769}$^\dagger$ \\
 & \scriptsize (0.024) & \scriptsize (0.024) & \scriptsize (0.020) & \scriptsize (0.019) & \scriptsize (0.016) & \scriptsize (0.009) & \scriptsize (0.008) & \scriptsize (0.006) & \scriptsize (0.004) \\[2pt]
Q-Simple (Qini) & 0.140 & 0.206$^\dagger$ & 0.378$^\dagger$ & \textbf{0.519}$^\dagger$ & 0.624$^\dagger$ & 0.681$^\dagger$ & 0.714$^\dagger$ & 0.766$^\dagger$ & 0.767$^\dagger$ \\
 & \scriptsize (0.026) & \scriptsize (0.021) & \scriptsize (0.025) & \scriptsize (0.021) & \scriptsize (0.013) & \scriptsize (0.009) & \scriptsize (0.009) & \scriptsize (0.006) & \scriptsize (0.004) \\[2pt]
\hline
\end{tabular}
\end{table}

\begin{table}[H]
\centering
\caption{Pointwise calibration error, as in Table~\ref{tab:app-syn-qini}
(geometric mean). Second line: share of runs whose pointwise calibration slope
lies in $[0.8, 1.25]$; $\infty$: at least one run with an infinite error. A
constant predictor scores about $1.42$. Best per column in bold;
$^\dagger$: as in Table~\ref{tab:app-syn-qini} (log error).}
\label{tab:app-syn-cal}
\footnotesize
\setlength{\tabcolsep}{3pt}
\begin{tabular}{l|rrrrrrrrr}
\hline
Learner & 50 & 100 & 200 & 400 & 800 & 1600 & 3200 & 6400 & 12800 \\
\hline
S & 2.61 & 2.11 & 1.82 & 1.61 & 1.44 & 1.33 & 1.25 & 1.20 & 1.16 \\
 & \scriptsize 0\% & \scriptsize 0\% & \scriptsize 0\% & \scriptsize 0\% & \scriptsize 0\% & \scriptsize 5\% & \scriptsize 77\% & \scriptsize 100\% & \scriptsize 100\% \\[2pt]
T & 9.42 & 8.01 & 5.30 & 3.77 & 2.59 & 2.02 & 1.58 & 1.37 & 1.25 \\
 & \scriptsize 0\% & \scriptsize 0\% & \scriptsize 0\% & \scriptsize 0\% & \scriptsize 0\% & \scriptsize 0\% & \scriptsize 0\% & \scriptsize 0\% & \scriptsize 0\% \\[2pt]
\hline
Q & 2.20 & 2.22 & 2.34 & 2.38 & 2.39 & 2.06 & 1.71 & 1.47 & 1.32 \\
 & \scriptsize 0\% & \scriptsize 0\% & \scriptsize 0\% & \scriptsize 0\% & \scriptsize 0\% & \scriptsize 0\% & \scriptsize 0\% & \scriptsize 0\% & \scriptsize 0\% \\[2pt]
Q-Simple & 1.78 & 1.93 & 2.16 & 2.30 & 2.39 & 2.06 & 1.72 & 1.47 & 1.32 \\
 & \scriptsize 0\% & \scriptsize 0\% & \scriptsize 0\% & \scriptsize 0\% & \scriptsize 0\% & \scriptsize 0\% & \scriptsize 0\% & \scriptsize 0\% & \scriptsize 0\% \\[2pt]
\hline
X & $10^{5}$ & $10^{4}$ & $10^{3}$ & 101.10 & 14.86 & 5.20 & 2.36 & 1.65 & 1.36 \\
 & \scriptsize 0\% & \scriptsize 0\% & \scriptsize 0\% & \scriptsize 0\% & \scriptsize 0\% & \scriptsize 0\% & \scriptsize 0\% & \scriptsize 0\% & \scriptsize 0\% \\[2pt]
R & $10^{5}$ & $10^{5}$ & $10^{4}$ & 225.38 & 28.35 & 7.48 & 3.13 & 1.93 & 1.48 \\
 & \scriptsize 0\% & \scriptsize 0\% & \scriptsize 0\% & \scriptsize 0\% & \scriptsize 0\% & \scriptsize 0\% & \scriptsize 0\% & \scriptsize 0\% & \scriptsize 0\% \\[2pt]
AIPW-DR & $10^{5}$ & $10^{5}$ & $10^{4}$ & 213.36 & 27.64 & 7.29 & 3.13 & 1.93 & 1.48 \\
 & \scriptsize 0\% & \scriptsize 0\% & \scriptsize 0\% & \scriptsize 0\% & \scriptsize 0\% & \scriptsize 0\% & \scriptsize 0\% & \scriptsize 0\% & \scriptsize 0\% \\[2pt]
\hline
DR-S & $10^{6}$ & $10^{6}$ & $10^{4}$ & $10^{3}$ & 91.27 & 374.32 & 3.50 & 2.22 & 1.68 \\
 & \scriptsize 0\% & \scriptsize 0\% & \scriptsize 0\% & \scriptsize 0\% & \scriptsize 0\% & \scriptsize 0\% & \scriptsize 0\% & \scriptsize 0\% & \scriptsize 0\% \\[2pt]
DR-T & $10^{7}$ & $10^{7}$ & $10^{6}$ & $10^{7}$ & $10^{6}$ & $10^{7}$ & $10^{5}$ & 6.10 & 2.58 \\
 & \scriptsize 0\% & \scriptsize 0\% & \scriptsize 0\% & \scriptsize 0\% & \scriptsize 0\% & \scriptsize 0\% & \scriptsize 0\% & \scriptsize 0\% & \scriptsize 0\% \\[2pt]
DR-Q & $10^{6}$ & $10^{6}$ & $10^{5}$ & $10^{7}$ & $10^{7}$ & $10^{7}$ & $10^{6}$ & 90.60 & 3.86 \\
 & \scriptsize 0\% & \scriptsize 0\% & \scriptsize 0\% & \scriptsize 0\% & \scriptsize 0\% & \scriptsize 0\% & \scriptsize 0\% & \scriptsize 0\% & \scriptsize 0\% \\[2pt]
DR-Q-Simple & 22.12 & $10^{4}$ & $10^{5}$ & $10^{6}$ & $10^{6}$ & $10^{5}$ & $10^{3}$ & 111.62 & 15.93 \\
 & \scriptsize 0\% & \scriptsize 0\% & \scriptsize 0\% & \scriptsize 0\% & \scriptsize 0\% & \scriptsize 0\% & \scriptsize 0\% & \scriptsize 0\% & \scriptsize 0\% \\[2pt]
\hline
DR-S log & $\infty$ & $\infty$ & $\infty$ & $\infty$ & $\infty$ & $\infty$ & $\infty$ & $\infty$ & $\infty$ \\
 & \scriptsize 0\% & \scriptsize 0\% & \scriptsize 0\% & \scriptsize 0\% & \scriptsize 0\% & \scriptsize 0\% & \scriptsize 0\% & \scriptsize 0\% & \scriptsize 0\% \\[2pt]
DR-T log & $\infty$ & $\infty$ & $\infty$ & $\infty$ & $\infty$ & $\infty$ & $\infty$ & $\infty$ & $\infty$ \\
 & \scriptsize 0\% & \scriptsize 0\% & \scriptsize 0\% & \scriptsize 0\% & \scriptsize 0\% & \scriptsize 0\% & \scriptsize 0\% & \scriptsize 0\% & \scriptsize 0\% \\[2pt]
DR-Q log & $\infty$ & $\infty$ & $\infty$ & $\infty$ & $\infty$ & $\infty$ & $\infty$ & $\infty$ & $\infty$ \\
 & \scriptsize 0\% & \scriptsize 0\% & \scriptsize 0\% & \scriptsize 0\% & \scriptsize 0\% & \scriptsize 0\% & \scriptsize 0\% & \scriptsize 0\% & \scriptsize 0\% \\[2pt]
DR-Q-Simple log & 1.73 & 1.94 & 2.27 & 2.58 & 2.96 & 2.56 & 1.94 & 1.57 & 1.35 \\
 & \scriptsize 2\% & \scriptsize 0\% & \scriptsize 0\% & \scriptsize 0\% & \scriptsize 0\% & \scriptsize 0\% & \scriptsize 0\% & \scriptsize 0\% & \scriptsize 0\% \\[2pt]
\hline
S (log-loss) & \textbf{1.44}$^\dagger$ & \textbf{1.44}$^\dagger$ & \textbf{1.40}$^\dagger$ & 1.36$^\dagger$ & 1.32$^\dagger$ & 1.26$^\dagger$ & 1.24$^\dagger$ & 1.20 & 1.14$^\dagger$ \\
 & \scriptsize 12\% & \scriptsize 0\% & \scriptsize 5\% & \scriptsize 10\% & \scriptsize 5\% & \scriptsize 8\% & \scriptsize 0\% & \scriptsize 7\% & \scriptsize 53\% \\[2pt]
S (Qini) & 1.50$^\dagger$ & 1.45$^\dagger$ & 1.40$^\dagger$ & \textbf{1.34}$^\dagger$ & 1.29$^\dagger$ & 1.28$^\dagger$ & 1.22$^\dagger$ & 1.16$^\dagger$ & 1.13$^\dagger$ \\
 & \scriptsize 8\% & \scriptsize 38\% & \scriptsize 30\% & \scriptsize 35\% & \scriptsize 42\% & \scriptsize 22\% & \scriptsize 47\% & \scriptsize 70\% & \scriptsize 83\% \\[2pt]
T (log-loss) & 1.67$^\dagger$ & 1.48$^\dagger$ & 1.41$^\dagger$ & 1.36$^\dagger$ & \textbf{1.25}$^\dagger$ & 1.21$^\dagger$ & \textbf{1.17}$^\dagger$ & 1.14$^\dagger$ & \textbf{1.12}$^\dagger$ \\
 & \scriptsize 2\% & \scriptsize 22\% & \scriptsize 38\% & \scriptsize 30\% & \scriptsize 85\% & \scriptsize 98\% & \scriptsize 97\% & \scriptsize 100\% & \scriptsize 100\% \\[2pt]
Q-Simple (log-loss) & 1.66$^\dagger$ & 1.53$^\dagger$ & 1.43$^\dagger$ & 1.35$^\dagger$ & 1.26$^\dagger$ & \textbf{1.20}$^\dagger$ & 1.17$^\dagger$ & 1.14$^\dagger$ & 1.12$^\dagger$ \\
 & \scriptsize 2\% & \scriptsize 8\% & \scriptsize 38\% & \scriptsize 45\% & \scriptsize 80\% & \scriptsize 82\% & \scriptsize 87\% & \scriptsize 100\% & \scriptsize 100\% \\[2pt]
Q-Simple (Qini) & 2.73$^\dagger$ & 1.74$^\dagger$ & 1.60$^\dagger$ & 1.37$^\dagger$ & 1.29$^\dagger$ & 1.23$^\dagger$ & 1.18$^\dagger$ & \textbf{1.14}$^\dagger$ & 1.14$^\dagger$ \\
 & \scriptsize 0\% & \scriptsize 12\% & \scriptsize 20\% & \scriptsize 18\% & \scriptsize 45\% & \scriptsize 78\% & \scriptsize 90\% & \scriptsize 97\% & \scriptsize 67\% \\[2pt]
\hline
\end{tabular}
\end{table}

\begin{figure}[H]
\centering
\figorpending{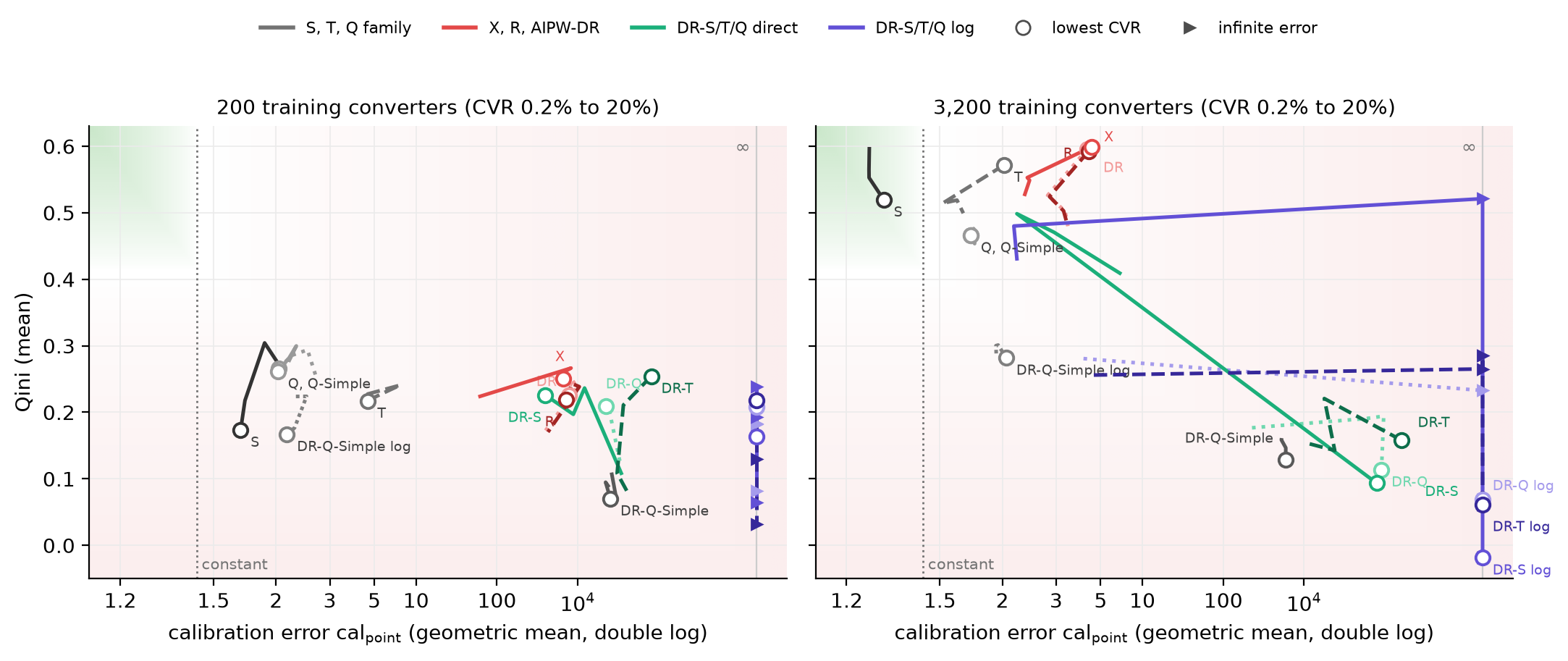}{\textwidth}
\caption{As Figure~\ref{fig:syn-untuned-v2}, for all untuned learners.}
\label{fig:syn-untuned-all}
\end{figure}

\paragraph{All untuned learners.}
Figure~\ref{fig:syn-untuned-all} repeats Figure~\ref{fig:syn-untuned-v2} for
every untuned learner.

\paragraph{Confounded assignment.}
Tables~\ref{tab:app-synthconf-qini} and~\ref{tab:app-synthconf-cal} report
every learner on the confounded variant of Section~\ref{sec:syn-conf-v2}.

\begin{table}[H]
\centering
\caption{Population Qini on the confounded synthetic data (CVR $2\%$, $e(x)$
estimated), by training converters (mean over $10$ seeds; standard error in
parentheses). Best per column in bold; $^\dagger$: tuned variant
significantly different from its untuned learner.}
\label{tab:app-synthconf-qini}
\footnotesize
\setlength{\tabcolsep}{3pt}
\begin{tabular}{l|rrr}
\hline
Learner & 800 & 3200 & 12800 \\
\hline
S & 0.408 & 0.542 & 0.659 \\
 & \scriptsize (0.021) & \scriptsize (0.019) & \scriptsize (0.012) \\[2pt]
T & 0.402 & 0.501 & 0.629 \\
 & \scriptsize (0.020) & \scriptsize (0.017) & \scriptsize (0.008) \\[2pt]
\hline
Q & 0.292 & 0.437 & 0.569 \\
 & \scriptsize (0.030) & \scriptsize (0.015) & \scriptsize (0.010) \\[2pt]
\hline
X & 0.390 & 0.513 & 0.659 \\
 & \scriptsize (0.020) & \scriptsize (0.018) & \scriptsize (0.009) \\[2pt]
R & 0.356 & 0.482 & 0.636 \\
 & \scriptsize (0.017) & \scriptsize (0.017) & \scriptsize (0.009) \\[2pt]
AIPW-DR & 0.358 & 0.482 & 0.638 \\
 & \scriptsize (0.017) & \scriptsize (0.017) & \scriptsize (0.009) \\[2pt]
\hline
DR-S & 0.296 & 0.483 & 0.655 \\
 & \scriptsize (0.030) & \scriptsize (0.030) & \scriptsize (0.012) \\[2pt]
DR-T & 0.196 & 0.164 & 0.139 \\
 & \scriptsize (0.035) & \scriptsize (0.042) & \scriptsize (0.025) \\[2pt]
DR-Q & 0.265 & 0.197 & 0.185 \\
 & \scriptsize (0.037) & \scriptsize (0.041) & \scriptsize (0.040) \\[2pt]
\hline
DR-S log & 0.254 & 0.504 & 0.653 \\
 & \scriptsize (0.034) & \scriptsize (0.020) & \scriptsize (0.010) \\[2pt]
DR-T log & 0.159 & 0.228 & 0.608 \\
 & \scriptsize (0.020) & \scriptsize (0.037) & \scriptsize (0.012) \\[2pt]
DR-Q log & 0.156 & 0.209 & 0.608 \\
 & \scriptsize (0.037) & \scriptsize (0.034) & \scriptsize (0.011) \\[2pt]
\hline
S (log-loss) & 0.516$^\dagger$ & 0.611$^\dagger$ & 0.651 \\
 & \scriptsize (0.019) & \scriptsize (0.016) & \scriptsize (0.013) \\[2pt]
S (Qini) & 0.551$^\dagger$ & \textbf{0.704}$^\dagger$ & \textbf{0.753}$^\dagger$ \\
 & \scriptsize (0.027) & \scriptsize (0.015) & \scriptsize (0.010) \\[2pt]
T (log-loss) & \textbf{0.695}$^\dagger$ & 0.690$^\dagger$ & 0.731$^\dagger$ \\
 & \scriptsize (0.017) & \scriptsize (0.013) & \scriptsize (0.010) \\[2pt]
Q (log-loss) & 0.597$^\dagger$ & 0.669$^\dagger$ & 0.753$^\dagger$ \\
 & \scriptsize (0.019) & \scriptsize (0.018) & \scriptsize (0.008) \\[2pt]
Q (Qini) & 0.415$^\dagger$ & 0.657$^\dagger$ & 0.735$^\dagger$ \\
 & \scriptsize (0.028) & \scriptsize (0.016) & \scriptsize (0.010) \\[2pt]
\hline
\end{tabular}
\end{table}

\begin{table}[H]
\centering
\caption{Pointwise calibration error on the confounded synthetic data
(geometric mean over $10$ seeds). Second line: share of seeds whose slope lies in
$[0.8, 1.25]$. Best per column in bold; $^\dagger$ as in
Table~\ref{tab:app-synthconf-qini}; $\infty$: at least one run with an infinite error; `--': no slope.}
\label{tab:app-synthconf-cal}
\footnotesize
\setlength{\tabcolsep}{3pt}
\begin{tabular}{l|rrr}
\hline
Learner & 800 & 3200 & 12800 \\
\hline
S & 1.41 & 1.25 & 1.16 \\
 & \scriptsize 0\% & \scriptsize 80\% & \scriptsize 100\% \\[2pt]
T & 2.76 & 1.58 & 1.26 \\
 & \scriptsize 0\% & \scriptsize 0\% & \scriptsize 0\% \\[2pt]
\hline
Q & 2.77 & 1.81 & 1.35 \\
 & \scriptsize 0\% & \scriptsize 0\% & \scriptsize 0\% \\[2pt]
\hline
X & 25.89 & 2.78 & 1.45 \\
 & \scriptsize 0\% & \scriptsize 0\% & \scriptsize 0\% \\[2pt]
R & 48.41 & 3.75 & 1.56 \\
 & \scriptsize 0\% & \scriptsize 0\% & \scriptsize 0\% \\[2pt]
AIPW-DR & 59.46 & 3.86 & 1.58 \\
 & \scriptsize 0\% & \scriptsize 0\% & \scriptsize 0\% \\[2pt]
\hline
DR-S & 10.94 & 2.54 & 1.51 \\
 & \scriptsize 0\% & \scriptsize 0\% & \scriptsize 0\% \\[2pt]
DR-T & $10^{7}$ & $10^{6}$ & 2.68 \\
 & \scriptsize 0\% & \scriptsize 0\% & \scriptsize 0\% \\[2pt]
DR-Q & $10^{6}$ & $10^{8}$ & 3.78 \\
 & \scriptsize 0\% & \scriptsize 0\% & \scriptsize 0\% \\[2pt]
\hline
DR-S log & $\infty$ & $\infty$ & $\infty$ \\
 & \scriptsize 0\% & \scriptsize 0\% & \scriptsize 0\% \\[2pt]
DR-T log & $\infty$ & $\infty$ & $\infty$ \\
 & \scriptsize -- & \scriptsize 0\% & \scriptsize 0\% \\[2pt]
DR-Q log & $\infty$ & $\infty$ & $\infty$ \\
 & \scriptsize -- & \scriptsize 0\% & \scriptsize 0\% \\[2pt]
\hline
S (log-loss) & 1.32$^\dagger$ & 1.22$^\dagger$ & 1.18$^\dagger$ \\
 & \scriptsize 0\% & \scriptsize 0\% & \scriptsize 0\% \\[2pt]
S (Qini) & 1.29$^\dagger$ & 1.26 & 1.13$^\dagger$ \\
 & \scriptsize 80\% & \scriptsize 0\% & \scriptsize 90\% \\[2pt]
T (log-loss) & 1.30$^\dagger$ & \textbf{1.17}$^\dagger$ & \textbf{1.13}$^\dagger$ \\
 & \scriptsize 80\% & \scriptsize 100\% & \scriptsize 100\% \\[2pt]
Q (log-loss) & \textbf{1.29}$^\dagger$ & 1.20$^\dagger$ & 1.14$^\dagger$ \\
 & \scriptsize 70\% & \scriptsize 100\% & \scriptsize 100\% \\[2pt]
Q (Qini) & 1.72$^\dagger$ & 1.22$^\dagger$ & 1.17$^\dagger$ \\
 & \scriptsize 0\% & \scriptsize 80\% & \scriptsize 40\% \\[2pt]
\hline
\end{tabular}
\end{table}

\section{Full Results on Randomized Data}
\label{app:real-tables}
Tables~\ref{tab:app-rct-qini} and~\ref{tab:app-rct-cal} report every learner
on every randomized dataset, at $12{,}800$ training converters or the whole
pool. A dagger marks a tuned variant whose paired difference
to its untuned learner is significant (Appendix~\ref{app:stest}). Two results
are not visible in the tables. The direct-scale augmentations underpredict the level on every randomized
dataset except DR-S on X5, by a median of about $2$ to $11$ log units, and on H(Conv) all of
them sit at the floor of $10^{-10}$, as on synthetic data. Unlike on
synthetic data, DR-Q-Simple on the log scale explodes on H(Conv), the
smallest dataset: its calibration error exceeds $10$ in every split
and $100$ in two thirds of them.

\begin{table}[H]
\centering
\caption{Count-based Qini on the randomized datasets, sorted by conversion
rate, at $12{,}800$ training converters or the whole pool (mean over $30$
seeds; corrected standard error in parentheses). Best per column in bold;
$^\dagger$: tuned variant significantly different from its untuned learner.}
\label{tab:app-rct-qini}
\footnotesize
\setlength{\tabcolsep}{3pt}
\begin{tabular}{l|rrrrrrr}
\hline
Learner & H(Conv) & Twins & Criteo & Lenta & H(Vis) & MegaFon & X5 \\
\hline
S & 0.689 & 0.082 & \textbf{0.527} & 0.023 & 0.433 & 1.251 & 0.049 \\
 & \scriptsize (0.271) & \scriptsize (0.020) & \scriptsize (0.051) & \scriptsize (0.013) & \scriptsize (0.091) & \scriptsize (0.028) & \scriptsize (0.008) \\[2pt]
T & 0.588 & 0.071 & 0.364 & 0.014 & 0.290 & 1.229 & 0.034 \\
 & \scriptsize (0.341) & \scriptsize (0.037) & \scriptsize (0.032) & \scriptsize (0.010) & \scriptsize (0.068) & \scriptsize (0.030) & \scriptsize (0.005) \\[2pt]
\hline
Q & 0.433 & 0.088 & 0.254 & 0.002 & 0.202 & 0.810 & 0.017 \\
 & \scriptsize (0.174) & \scriptsize (0.039) & \scriptsize (0.027) & \scriptsize (0.011) & \scriptsize (0.055) & \scriptsize (0.023) & \scriptsize (0.006) \\[2pt]
Q-Simple & 0.491 & 0.086 & 0.286 & $-$0.001 & 0.193 & 0.828 & 0.011 \\
 & \scriptsize (0.198) & \scriptsize (0.039) & \scriptsize (0.026) & \scriptsize (0.011) & \scriptsize (0.050) & \scriptsize (0.024) & \scriptsize (0.007) \\[2pt]
\hline
X & 0.584 & 0.064 & 0.420 & 0.021 & 0.338 & 1.247 & 0.040 \\
 & \scriptsize (0.259) & \scriptsize (0.033) & \scriptsize (0.047) & \scriptsize (0.010) & \scriptsize (0.086) & \scriptsize (0.028) & \scriptsize (0.006) \\[2pt]
R & 0.699 & 0.065 & 0.428 & 0.015 & 0.353 & 1.155 & 0.032 \\
 & \scriptsize (0.306) & \scriptsize (0.029) & \scriptsize (0.048) & \scriptsize (0.013) & \scriptsize (0.089) & \scriptsize (0.029) & \scriptsize (0.006) \\[2pt]
AIPW-DR & \textbf{0.798} & 0.040 & 0.442 & 0.011 & 0.352 & 1.159 & 0.034 \\
 & \scriptsize (0.515) & \scriptsize (0.028) & \scriptsize (0.051) & \scriptsize (0.012) & \scriptsize (0.074) & \scriptsize (0.031) & \scriptsize (0.006) \\[2pt]
\hline
DR-S & 0.480 & 0.012 & 0.111 & 0.022 & 0.352 & 0.695 & 0.033 \\
 & \scriptsize (0.265) & \scriptsize (0.048) & \scriptsize (0.021) & \scriptsize (0.019) & \scriptsize (0.097) & \scriptsize (0.016) & \scriptsize (0.006) \\[2pt]
DR-T & $-$0.050 & 0.030 & 0.072 & 0.009 & 0.270 & 0.502 & 0.030 \\
 & \scriptsize (0.159) & \scriptsize (0.063) & \scriptsize (0.019) & \scriptsize (0.012) & \scriptsize (0.114) & \scriptsize (0.024) & \scriptsize (0.008) \\[2pt]
DR-Q & 0.535 & 0.009 & 0.148 & 0.007 & 0.265 & 0.517 & 0.035 \\
 & \scriptsize (0.327) & \scriptsize (0.044) & \scriptsize (0.018) & \scriptsize (0.015) & \scriptsize (0.106) & \scriptsize (0.025) & \scriptsize (0.008) \\[2pt]
DR-Q-Simple & 0.490 & 0.000 & 0.117 & $-$0.004 & 0.232 & 0.553 & 0.012 \\
 & \scriptsize (0.301) & \scriptsize (0.040) & \scriptsize (0.028) & \scriptsize (0.010) & \scriptsize (0.073) & \scriptsize (0.024) & \scriptsize (0.006) \\[2pt]
\hline
DR-S log & 0.567 & 0.024 & 0.115 & 0.025 & 0.343 & 0.691 & 0.034 \\
 & \scriptsize (0.266) & \scriptsize (0.061) & \scriptsize (0.019) & \scriptsize (0.016) & \scriptsize (0.099) & \scriptsize (0.023) & \scriptsize (0.006) \\[2pt]
DR-T log & 0.163 & 0.025 & 0.123 & 0.022 & 0.343 & 0.558 & 0.036 \\
 & \scriptsize (0.256) & \scriptsize (0.045) & \scriptsize (0.015) & \scriptsize (0.018) & \scriptsize (0.095) & \scriptsize (0.017) & \scriptsize (0.007) \\[2pt]
DR-Q log & 0.495 & 0.004 & 0.166 & 0.014 & 0.307 & 0.590 & 0.035 \\
 & \scriptsize (0.289) & \scriptsize (0.043) & \scriptsize (0.022) & \scriptsize (0.016) & \scriptsize (0.085) & \scriptsize (0.020) & \scriptsize (0.007) \\[2pt]
DR-Q-Simple log & 0.577 & 0.076 & 0.187 & 0.005 & 0.240 & 0.812 & 0.011 \\
 & \scriptsize (0.327) & \scriptsize (0.071) & \scriptsize (0.020) & \scriptsize (0.011) & \scriptsize (0.059) & \scriptsize (0.018) & \scriptsize (0.006) \\[2pt]
\hline
S (log-loss) & 0.466 & 0.094 & 0.252$^\dagger$ & 0.016 & 0.477 & 1.363$^\dagger$ & 0.047 \\
 & \scriptsize (0.272) & \scriptsize (0.019) & \scriptsize (0.037) & \scriptsize (0.012) & \scriptsize (0.092) & \scriptsize (0.029) & \scriptsize (0.009) \\[2pt]
S (Qini) & 0.498 & \textbf{0.103} & 0.407$^\dagger$ & \textbf{0.035} & 0.498 & \textbf{1.418}$^\dagger$ & \textbf{0.071}$^\dagger$ \\
 & \scriptsize (0.232) & \scriptsize (0.026) & \scriptsize (0.031) & \scriptsize (0.021) & \scriptsize (0.086) & \scriptsize (0.029) & \scriptsize (0.010) \\[2pt]
T (log-loss) & 0.373 & 0.093 & 0.240$^\dagger$ & 0.025 & 0.457$^\dagger$ & 1.380$^\dagger$ & 0.048 \\
 & \scriptsize (0.284) & \scriptsize (0.031) & \scriptsize (0.025) & \scriptsize (0.011) & \scriptsize (0.081) & \scriptsize (0.026) & \scriptsize (0.008) \\[2pt]
Q-Simple (log-loss) & 0.384 & 0.082 & 0.355$^\dagger$ & 0.024 & 0.500$^\dagger$ & 0.978$^\dagger$ & 0.036$^\dagger$ \\
 & \scriptsize (0.189) & \scriptsize (0.018) & \scriptsize (0.033) & \scriptsize (0.016) & \scriptsize (0.118) & \scriptsize (0.021) & \scriptsize (0.012) \\[2pt]
Q-Simple (Qini) & 0.580 & 0.076 & 0.405$^\dagger$ & 0.018 & \textbf{0.564}$^\dagger$ & 1.007$^\dagger$ & 0.051$^\dagger$ \\
 & \scriptsize (0.242) & \scriptsize (0.019) & \scriptsize (0.045) & \scriptsize (0.013) & \scriptsize (0.120) & \scriptsize (0.021) & \scriptsize (0.013) \\[2pt]
\hline
\end{tabular}
\end{table}

\begin{table}[H]
\centering
\caption{Binned calibration error on the randomized datasets, as in
Table~\ref{tab:app-rct-qini} (geometric mean over $30$ seeds). Second line:
share of seeds whose binned calibration slope lies in $[0.8, 1.25]$; ``--'':
no slope (single bucket). Best per column in bold; $^\dagger$: tuned variant
significantly different from its untuned learner (log error).}
\label{tab:app-rct-cal}
\footnotesize
\setlength{\tabcolsep}{3pt}
\begin{tabular}{l|rrrrrrr}
\hline
Learner & H(Conv) & Twins & Criteo & Lenta & H(Vis) & MegaFon & X5 \\
\hline
S & 2.72 & 1.41 & 1.31 & 1.07 & 1.22 & 1.19 & 1.05 \\
 & \scriptsize -- & \scriptsize 23\% & \scriptsize 40\% & \scriptsize 17\% & \scriptsize 7\% & \scriptsize 0\% & \scriptsize 0\% \\[2pt]
T & 31.14 & 2.17 & 2.05 & 1.36 & 1.49 & 1.08 & 1.18 \\
 & \scriptsize -- & \scriptsize 0\% & \scriptsize 0\% & \scriptsize 0\% & \scriptsize 0\% & \scriptsize 100\% & \scriptsize 0\% \\[2pt]
\hline
Q & 9.14 & 2.12 & 1.43 & 1.33 & 1.51 & 1.09 & 1.26 \\
 & \scriptsize -- & \scriptsize 0\% & \scriptsize 0\% & \scriptsize 0\% & \scriptsize 0\% & \scriptsize 100\% & \scriptsize 0\% \\[2pt]
Q-Simple & 9.17 & 2.13 & 1.44 & 1.32 & 1.52 & 1.07 & 1.28 \\
 & \scriptsize -- & \scriptsize 0\% & \scriptsize 0\% & \scriptsize 0\% & \scriptsize 0\% & \scriptsize 100\% & \scriptsize 0\% \\[2pt]
\hline
X & 33.52 & 412.63 & 9.72 & 1.22 & 1.39 & 1.60 & 1.12 \\
 & \scriptsize -- & \scriptsize 0\% & \scriptsize 0\% & \scriptsize 0\% & \scriptsize 0\% & \scriptsize 10\% & \scriptsize 0\% \\[2pt]
R & 32.57 & $10^{3}$ & 3.92 & 1.26 & 1.44 & 9.43 & 1.18 \\
 & \scriptsize -- & \scriptsize 0\% & \scriptsize 0\% & \scriptsize 0\% & \scriptsize 0\% & \scriptsize 0\% & \scriptsize 0\% \\[2pt]
AIPW-DR & 32.92 & $10^{3}$ & 2.97 & 1.25 & 1.44 & 9.46 & 1.18 \\
 & \scriptsize -- & \scriptsize 0\% & \scriptsize 0\% & \scriptsize 0\% & \scriptsize 0\% & \scriptsize 0\% & \scriptsize 0\% \\[2pt]
\hline
DR-S & $10^{10}$ & 442.64 & 22.09 & 12.81 & 13.52 & 7.98 & 1.23 \\
 & \scriptsize -- & \scriptsize 0\% & \scriptsize 0\% & \scriptsize 0\% & \scriptsize 0\% & \scriptsize 0\% & \scriptsize 0\% \\[2pt]
DR-T & $10^{10}$ & $10^{3}$ & $10^{6}$ & 23.08 & 16.92 & 12.85 & 11.17 \\
 & \scriptsize -- & \scriptsize 0\% & \scriptsize 0\% & \scriptsize 0\% & \scriptsize 0\% & \scriptsize 0\% & \scriptsize 0\% \\[2pt]
DR-Q & $10^{10}$ & 590.46 & 38.32 & 15.81 & 17.78 & 11.90 & 11.55 \\
 & \scriptsize -- & \scriptsize 0\% & \scriptsize 0\% & \scriptsize 0\% & \scriptsize 0\% & \scriptsize 0\% & \scriptsize 0\% \\[2pt]
DR-Q-Simple & $10^{10}$ & $10^{5}$ & 50.12 & 13.25 & 181.48 & 12.00 & 10.03 \\
 & \scriptsize -- & \scriptsize 0\% & \scriptsize 0\% & \scriptsize 0\% & \scriptsize 0\% & \scriptsize 0\% & \scriptsize 0\% \\[2pt]
\hline
DR-S log & $\infty$ & $10^{12}$ & $\infty$ & 1.97 & 1.49 & 1.10 & 1.22 \\
 & \scriptsize -- & \scriptsize 0\% & \scriptsize 0\% & \scriptsize 0\% & \scriptsize 0\% & \scriptsize 100\% & \scriptsize 0\% \\[2pt]
DR-T log & $\infty$ & $\infty$ & $\infty$ & 2.57 & 1.66 & 1.18 & 1.28 \\
 & \scriptsize -- & \scriptsize 0\% & \scriptsize 0\% & \scriptsize 0\% & \scriptsize 0\% & \scriptsize 3\% & \scriptsize 0\% \\[2pt]
DR-Q log & $\infty$ & $\infty$ & $\infty$ & 8.58 & 1.66 & 1.21 & 1.27 \\
 & \scriptsize -- & \scriptsize 0\% & \scriptsize 0\% & \scriptsize 0\% & \scriptsize 0\% & \scriptsize 10\% & \scriptsize 0\% \\[2pt]
DR-Q-Simple log & 627.59 & 2.99 & 2.32 & 1.35 & 1.64 & 1.08 & 1.27 \\
 & \scriptsize -- & \scriptsize 0\% & \scriptsize 0\% & \scriptsize 0\% & \scriptsize 0\% & \scriptsize 100\% & \scriptsize 0\% \\[2pt]
\hline
S (log-loss) & 1.25$^\dagger$ & 1.39 & 1.13 & 1.08 & \textbf{1.15}$^\dagger$ & 1.27$^\dagger$ & \textbf{1.02}$^\dagger$ \\
 & \scriptsize -- & \scriptsize 13\% & \scriptsize 23\% & \scriptsize 0\% & \scriptsize 50\% & \scriptsize 0\% & \scriptsize 23\% \\[2pt]
S (Qini) & \textbf{1.23}$^\dagger$ & 1.39 & 1.42 & 1.08 & 1.16 & 1.18 & 1.02$^\dagger$ \\
 & \scriptsize -- & \scriptsize 17\% & \scriptsize 0\% & \scriptsize 17\% & \scriptsize 43\% & \scriptsize 0\% & \scriptsize 47\% \\[2pt]
T (log-loss) & 1.26$^\dagger$ & 1.38$^\dagger$ & 1.20$^\dagger$ & 1.16$^\dagger$ & 1.16$^\dagger$ & 1.08 & 1.06$^\dagger$ \\
 & \scriptsize -- & \scriptsize 7\% & \scriptsize 0\% & \scriptsize 0\% & \scriptsize 33\% & \scriptsize 100\% & \scriptsize 0\% \\[2pt]
Q-Simple (log-loss) & 1.26$^\dagger$ & \textbf{1.34}$^\dagger$ & \textbf{1.10}$^\dagger$ & \textbf{1.05}$^\dagger$ & 1.16$^\dagger$ & 1.08 & 1.03$^\dagger$ \\
 & \scriptsize -- & \scriptsize 20\% & \scriptsize 77\% & \scriptsize 20\% & \scriptsize 37\% & \scriptsize 100\% & \scriptsize 17\% \\[2pt]
Q-Simple (Qini) & 1.57$^\dagger$ & 1.36$^\dagger$ & 1.13$^\dagger$ & 1.10$^\dagger$ & 1.17$^\dagger$ & \textbf{1.06} & 1.03$^\dagger$ \\
 & \scriptsize -- & \scriptsize 3\% & \scriptsize 3\% & \scriptsize 0\% & \scriptsize 50\% & \scriptsize 100\% & \scriptsize 17\% \\[2pt]
\hline
\end{tabular}
\end{table}

\section{Observational Data}
\label{app:obs}
On the four observational datasets (Table~\ref{tab:obs-v2}) the propensity
must be estimated, so the Q-Learner takes the Q-Simple-Learner's place, and
all metrics are IPW-weighted. The datasets hold too little data for the
heterogeneity they carry. Cattaneo and NHEFS have only about $230$ and $250$
training converters, below the few hundred from which the learners rank
reliably on synthetic data. JTPA and RHC have about $3700$ and $3000$, but so
little heterogeneity that this is not enough either. Hardly any ranking
difference is significant, and the spread can rarely be resolved
(Tables~\ref{tab:app-obs-qini} and~\ref{tab:app-obs-cal}). One result stands
out: the tuned Q-Learner is less well calibrated than the tuned S-Learner,
significantly on two of the four datasets. Untuned, it is off in level by a
factor of seven on the two small datasets, since an untruncated $\hat e$
enters $\hat\tau$ directly. On the confounded synthetic data, the tuned
Q-Learner calibrates as well as the S-Learner
(Section~\ref{sec:syn-conf-v2}), which points to the small samples rather
than the estimated propensity. The metrics use self-normalized inverse
propensity weights $1/\hat e$ and $1/(1-\hat e)$, with $\hat e$ from the same
LightGBM propensity model the learners use. On Cattaneo and NHEFS, with about
$230$ and $250$ training converters and about $36$ and $44$ control converters
in a test set, the binned calibration has a single bucket: it measures the
level of $\hat\tau$ only, and there is no slope.

\begin{table}[H]
\centering
\caption{Observational datasets, sorted by conversion rate. Propensity must be
estimated. $^\dagger$Treatment is actual participation, not the randomized
offer, and therefore self-selected.}
\label{tab:obs-v2}
\setlength{\tabcolsep}{6pt}
\begin{tabular}{lrrrl}
\toprule
Dataset & $n$ & CVR & Treat. & Domain \\
\midrule
Cattaneo \citep{cattaneo2010} & 4.6k & 6.0\% & 18.6\% & Birthweight \\
NHEFS \citep{hernan2020} & 1.6k & 19.5\% & 26.3\% & Smoking \\
JTPA$^\dagger$ \citep{abadie2002} & 9.2k & 50.0\% & 44.0\% & Job training \\
RHC \citep{connors1996rhc} & 5.7k & 64.9\% & 38.1\% & Medical \\
\bottomrule
\end{tabular}
\end{table}

\begin{table}[H]
\centering
\caption{IPW-weighted Qini on the observational datasets, sorted by
conversion rate (mean over $30$ random splits; corrected standard error in
parentheses). Best per column in bold; $^\dagger$: tuned variant
significantly different from its untuned learner.}
\label{tab:app-obs-qini}
\footnotesize
\setlength{\tabcolsep}{3pt}
\begin{tabular}{l|rrrr}
\hline
Learner & Cattaneo & NHEFS & JTPA & RHC \\
\hline
S & 1.156 & \textbf{0.249} & 0.004 & 0.017 \\
 & \scriptsize (1.183) & \scriptsize (0.279) & \scriptsize (0.031) & \scriptsize (0.068) \\[2pt]
T & 0.746 & 0.184 & $-$0.014 & $-$0.022 \\
 & \scriptsize (0.502) & \scriptsize (0.305) & \scriptsize (0.030) & \scriptsize (0.077) \\[2pt]
\hline
Q & 0.884 & 0.098 & $-$0.009 & 0.018 \\
 & \scriptsize (0.658) & \scriptsize (0.159) & \scriptsize (0.031) & \scriptsize (0.043) \\[2pt]
\hline
X & 1.079 & 0.091 & $-$0.019 & $-$0.020 \\
 & \scriptsize (0.804) & \scriptsize (0.192) & \scriptsize (0.032) & \scriptsize (0.066) \\[2pt]
R & 1.391 & 0.158 & $-$0.021 & $-$0.010 \\
 & \scriptsize (1.162) & \scriptsize (0.215) & \scriptsize (0.030) & \scriptsize (0.065) \\[2pt]
AIPW-DR & 0.762 & 0.140 & $-$0.031 & 0.037 \\
 & \scriptsize (0.751) & \scriptsize (0.238) & \scriptsize (0.029) & \scriptsize (0.074) \\[2pt]
\hline
DR-S & 0.232 & 0.148 & $-$0.014 & $-$0.001 \\
 & \scriptsize (0.664) & \scriptsize (0.237) & \scriptsize (0.033) & \scriptsize (0.048) \\[2pt]
DR-T & 0.896 & $-$0.013 & $-$0.010 & $-$0.007 \\
 & \scriptsize (1.012) & \scriptsize (0.141) & \scriptsize (0.032) & \scriptsize (0.043) \\[2pt]
DR-Q & 0.732 & 0.086 & 0.010 & 0.008 \\
 & \scriptsize (0.782) & \scriptsize (0.156) & \scriptsize (0.033) & \scriptsize (0.054) \\[2pt]
\hline
DR-S log & 0.148 & 0.167 & $-$0.019 & 0.005 \\
 & \scriptsize (0.754) & \scriptsize (0.279) & \scriptsize (0.029) & \scriptsize (0.056) \\[2pt]
DR-T log & $-$0.266 & 0.117 & $-$0.012 & 0.027 \\
 & \scriptsize (0.552) & \scriptsize (0.214) & \scriptsize (0.036) & \scriptsize (0.045) \\[2pt]
DR-Q log & $-$0.416 & 0.028 & $-$0.005 & 0.019 \\
 & \scriptsize (0.626) & \scriptsize (0.137) & \scriptsize (0.032) & \scriptsize (0.047) \\[2pt]
\hline
S (log-loss) & 0.777 & 0.114 & 0.078$^\dagger$ & 0.043 \\
 & \scriptsize (0.722) & \scriptsize (0.132) & \scriptsize (0.032) & \scriptsize (0.055) \\[2pt]
S (Qini) & 1.157 & 0.165 & \textbf{0.136} & \textbf{0.060} \\
 & \scriptsize (0.894) & \scriptsize (0.200) & \scriptsize (0.066) & \scriptsize (0.068) \\[2pt]
T (log-loss) & 1.640 & 0.035 & 0.051$^\dagger$ & $-$0.002 \\
 & \scriptsize (0.909) & \scriptsize (0.201) & \scriptsize (0.035) & \scriptsize (0.050) \\[2pt]
Q (log-loss) & 1.146 & 0.242 & 0.095$^\dagger$ & 0.011 \\
 & \scriptsize (0.576) & \scriptsize (0.207) & \scriptsize (0.043) & \scriptsize (0.040) \\[2pt]
Q (Qini) & \textbf{1.836} & 0.207 & 0.134$^\dagger$ & 0.046 \\
 & \scriptsize (0.902) & \scriptsize (0.200) & \scriptsize (0.043) & \scriptsize (0.040) \\[2pt]
\hline
\end{tabular}
\end{table}

\begin{table}[H]
\centering
\caption{IPW-weighted binned calibration error on the observational datasets
(geometric mean over $30$ random splits). Second line: share of splits whose
slope lies in $[0.8, 1.25]$; ``--'': no slope, since Cattaneo and NHEFS have a
single calibration bucket (level only). Best per column in bold; $^\dagger$ as
in Table~\ref{tab:app-obs-qini}.}
\label{tab:app-obs-cal}
\footnotesize
\setlength{\tabcolsep}{3pt}
\begin{tabular}{l|rrrr}
\hline
Learner & Cattaneo & NHEFS & JTPA & RHC \\
\hline
S & 2.55 & 1.71 & 1.21 & \textbf{1.20} \\
 & \scriptsize -- & \scriptsize -- & \scriptsize 0\% & \scriptsize 13\% \\[2pt]
T & 23.28 & 26.74 & 1.38 & 1.52 \\
 & \scriptsize -- & \scriptsize -- & \scriptsize 0\% & \scriptsize 0\% \\[2pt]
\hline
Q & 7.52 & 6.66 & 1.43 & 1.84 \\
 & \scriptsize -- & \scriptsize -- & \scriptsize 0\% & \scriptsize 0\% \\[2pt]
\hline
X & 42.58 & 103.97 & 1.33 & 1.44 \\
 & \scriptsize -- & \scriptsize -- & \scriptsize 0\% & \scriptsize 0\% \\[2pt]
R & 45.21 & 200.01 & 1.42 & 8.86 \\
 & \scriptsize -- & \scriptsize -- & \scriptsize 0\% & \scriptsize 0\% \\[2pt]
AIPW-DR & 528.44 & $10^{4}$ & 1.46 & 15.89 \\
 & \scriptsize -- & \scriptsize -- & \scriptsize 0\% & \scriptsize 0\% \\[2pt]
\hline
DR-S & $10^{6}$ & 148.53 & 12.18 & 32.85 \\
 & \scriptsize -- & \scriptsize -- & \scriptsize 0\% & \scriptsize 0\% \\[2pt]
DR-T & $10^{10}$ & $10^{10}$ & 15.66 & $10^{3}$ \\
 & \scriptsize -- & \scriptsize -- & \scriptsize 0\% & \scriptsize 0\% \\[2pt]
DR-Q & $10^{10}$ & $10^{9}$ & 15.91 & $10^{4}$ \\
 & \scriptsize -- & \scriptsize -- & \scriptsize 0\% & \scriptsize 0\% \\[2pt]
\hline
DR-S log & $\infty$ & $\infty$ & 1.54 & 3.13 \\
 & \scriptsize -- & \scriptsize -- & \scriptsize 0\% & \scriptsize 0\% \\[2pt]
DR-T log & $\infty$ & $\infty$ & 1.62 & 16.25 \\
 & \scriptsize -- & \scriptsize -- & \scriptsize 0\% & \scriptsize 0\% \\[2pt]
DR-Q log & $\infty$ & $\infty$ & 1.56 & 6.11 \\
 & \scriptsize -- & \scriptsize -- & \scriptsize 0\% & \scriptsize 0\% \\[2pt]
\hline
S (log-loss) & \textbf{1.61} & \textbf{1.61} & \textbf{1.14}$^\dagger$ & 1.21 \\
 & \scriptsize -- & \scriptsize -- & \scriptsize 40\% & \scriptsize 7\% \\[2pt]
S (Qini) & 1.85 & 1.63 & 1.21 & 1.21 \\
 & \scriptsize -- & \scriptsize -- & \scriptsize 0\% & \scriptsize 17\% \\[2pt]
T (log-loss) & 1.98$^\dagger$ & 1.93$^\dagger$ & 1.15$^\dagger$ & 1.26$^\dagger$ \\
 & \scriptsize -- & \scriptsize -- & \scriptsize 7\% & \scriptsize 3\% \\[2pt]
Q (log-loss) & 2.12$^\dagger$ & 2.07$^\dagger$ & 1.16$^\dagger$ & 1.43$^\dagger$ \\
 & \scriptsize -- & \scriptsize -- & \scriptsize 3\% & \scriptsize 0\% \\[2pt]
Q (Qini) & 2.13$^\dagger$ & 7.21 & 1.17$^\dagger$ & 1.46$^\dagger$ \\
 & \scriptsize -- & \scriptsize -- & \scriptsize 7\% & \scriptsize 0\% \\[2pt]
\hline
\end{tabular}
\end{table}

\end{document}